\documentclass[runningheads]{llncs}

\usepackage{graphicx}
\usepackage{comment}
\usepackage{amsmath,amssymb}
\usepackage{color}
\usepackage{url}
\usepackage{hyperref}
\usepackage[dvipsnames]{xcolor}
\usepackage{oplotsymbl}
\usepackage{placeins}

\usepackage{booktabs}
\usepackage{caption}
\usepackage{subcaption}
\usepackage{multirow}
\usepackage{wrapfig}

\usepackage{orcidlink}

\newif\ifreview
\reviewfalse

\ifreview
	\usepackage{lineno}

	\linenumbers
\fi

\begin{document}


\def\SubNumber{}

\def\GCPRTrack{Main Track}

\title{An Evaluation of Zero-Cost Proxies -\\ from Neural Architecture Performance Prediction to Model Robustness}

\ifreview
\else
	\titlerunning{An Evaluation of Zero-Cost Proxies}

	\author{Jovita Lukasik\inst{1}\orcidlink{0000-0003-4243-9188} 
    \and
	Michael Moeller \inst{1}\orcidlink{0000-0002-0492-6527} 
 \and
	Margret Keuper \inst{1,2}\orcidlink{0000-0002-8437-7993}
	}
	\authorrunning{J. Lukasik et al.}
	
	\institute{University of Siegen \\
	\email{\{jovita.lukasik,michael.moeller,margret.keuper\}@uni-siegen.de}\\
 \and Max Planck Institute for Informatics, Saarland Informatics Campus 
 }
\fi

\maketitle              

\begin{abstract}
Zero-cost proxies are nowadays frequently studied and used to search for neural architectures. They show an impressive ability to predict the performance of architectures by making use of their untrained weights. These techniques allow for immense search speed-ups. So far the joint search for well-performing and robust architectures has received much less attention in the field of NAS. Therefore, the main focus of zero-cost proxies is the clean accuracy of architectures, whereas the model robustness should play an evenly important part. In this paper, we analyze the ability of common zero-cost proxies to serve as performance predictors for robustness in the popular NAS-Bench-201 search space. We are interested in the single prediction task for robustness and the joint multi-objective of clean and robust accuracy. We further analyze the feature importance of the proxies and show that predicting the robustness makes the prediction task from existing zero-cost proxies more challenging. As a result, the joint consideration of several proxies becomes necessary to predict a model's robustness while the clean accuracy can be regressed from a single such feature.
\end{abstract}
\section{Introduction}
Neural Architecture Search (NAS) \cite{2019NASSurvey,White23survey} seeks to automate the process of designing high-performing neural networks. This research field has gained immense popularity in the last years, with only a few NAS papers in 2016 to almost 700 in 2022 \cite{White23survey}. The ability of NAS to find high-performing architectures, capable of outperforming hand-designed ones, notably on image classification \cite{2017ReinforcementNAS}, makes this research field highly valuable. However, classical approaches like reinforcement learning \cite{2017ReinforcementNAS,2018ReinforcementNAS} or evolutionary algorithms \cite{2019EvolutionaryNAS} are expensive, which led to a focus-shift towards improving the search efficiency. Therefore different approaches such as one-shot methods \cite{2018DARTS} or performance prediction methods \cite{2021HowPP} were introduced. Recently, zero-cost proxies (ZCP) \cite{MellorTSC21}, as part of performance prediction strategies, were developed. These proxies build upon fast computations, mostly in one single forward and backward pass on an untrained model, and attempt to predict the accuracy that the underlying architecture will have after training. 

In recent years, several zero-cost proxies were introduced \cite{MellorTSC21,AbdelfattahMDL21},  including simple architectural baselines as FLOPS or the number of parameters. NAS-Bench-Suite-Zero \cite{KrishnakumarWZT22} provides a more in-depth analysis and evaluates 13 different ZCPs on 28 different tasks to show the effectiveness as a  performance prediction technique. 

So far, the main focus of NAS research was the resulting performance of the architectures on a downstream task. Another important aspect of networks, namely their robustness, has been less addressed in NAS so far. Most works targeting both high accuracy and a robust network rely on one-shot methods \cite{Hosseini21,Mok21}. However, using only ZCPs as a performance prediction model for the multi-objective has not been addressed. The search for architectures that are robust against adversarial attacks is especially important for computer vision and image classification since networks can be easily fooled when the input data is changed using slight perturbations that are even invisible to the human eye. This can lead to networks making false predictions with high confidence. 

This aspect is particularly important in the context of NAS and ZCPs 
because the search for robust architectures is significantly more expensive: an the architecture's robustness is to be evaluated on trained architectures against different adversarial attacks \cite{fgsm,pgd,croce2020autoattack}.

In this paper, we therefore address the question: How transferable are the accuracy-focused ZCPs to the architecture's robustness? A high-performing architecture is not necessarily robust. Therefore, we analyze which ZCPs perform well for predicting clean accuracy, which are good at predicting robustness, and which do well at both. 

For our evaluation, we leverage the recently published robustness dataset \cite{Jung2023} which allows for easily accessible robustness evaluations on an established NAS search space \cite{2020NB201}. 
Since every ZCP provides a low-dimensional (scalar) measure per architecture, we understand each ZCP as a feature dimension in a concatenated feature vector and employ random forest regression as a predictor of clean and robust accuracy. This facilitates not only the evaluation of the performance or correlation of the different measures with the prediction target but also to gain direct access to every proxy's feature importance compared with all others.

Our evaluation of all ZCPs from NAS-Bench-Suite-Zero allows us to make the following observations:
\begin{itemize}
    \item While the correlation of every single ZCP with the target is not very strong, the random forest regression allows predicting the clean accuracy with very good and the robust accuracy with good precision.  
    \item When analyzing the feature importance, ZCPs using Jacobian-based information generally carry the most employed information.
    \item The feature importance distribution shows that clean accuracies can be predicted from one or few ZCPs while regressors trained to predict the robust accuracy tend to rely on \emph{all} available information.
\end{itemize}

\section{Related Work}

\subsection{Zero-Cost Proxies for NAS}
NAS is the process to automate the design of neural architectures with the goal to find a high-performing architecture on a particular dataset. In the last few years, this research field gained immense popularity and is able to surpass hand-designed neural architectures on different tasks, especially on image classification. See \cite{White23survey} for a survey.   
For fast search and evaluation of found architectures, many NAS methods make use of performance prediction techniques via surrogate models. The surrogate model predicts the performance of an architecture, without the need of training the architecture, with the goal to keep the query amount (i.e.~the amount necessary training data to train the surrogate model) low. Each query means one full training of the architecture, therefore, successful performance prediction-based NAS methods are able to use only a few queries. \cite{2020NP,2020BONAS,2021BANANAS,2021WeakNAS,Lukasik2022} show improved results using surrogate models keeping the query amount low. 
However, these methods use the validation or test accuracy of the architecture as a target and require high computation time. In order to predict the performance of an architecture without full training, zero-cost proxies (ZCPs), measured on untrained architectures, can be used \cite{MellorTSC21}. The idea is that these ZCPs, which often require only one forward and one backward pass on a single mini-batch, are somehow correlated with the resulting performance of the architecture after full training.  
\cite{MellorTSC21} originally used ZCPs for NAS by analyzing the linear regions and how well they are separated. 
In contrast, \cite{AbdelfattahMDL21} uses pruning-at-initialization techniques (\cite{LeeAT19,WangZG20,TanakaKYG20}) as ZCPs. The best performing ZCP in \cite{AbdelfattahMDL21} \texttt{synflow} is data-independent, which does not consider the input data for the proxy calculation.
Another data-independent architecture was proposed by \cite{LinWSCS00021}. 
Other approaches use the neural tangent kernel for faster architecture search \cite{2021Im4GPU}.

The recent benchmark NAS-Bench-suite zero \cite{KrishnakumarWZT22} compares different ZCPs on different NAS search spaces and shows how they can be integrated into different NAS frameworks. 
Other works include ZCPs into Bayesian optimization NAS approaches (as for example \cite{Shen2021,2021HowPP}) and one-shot architecture search \cite{Xiang2021}. In contrast, this paper evaluates how well ZCPs can predict the robust accuracy of a model under adversarial attacks, and demonstrates (surprising) success in combining different ZCP features in a random-forest classifier.

\subsection{Robustness in NAS}
Compared to searching for an architecture with the single objective of having a high performance, including the architecture's robustness results in an even more challenging task that requires a multi-objective search. 
Recent works that search for both high-performing and robust architectures combine both objectives in one-shot search approaches \cite{Guo2020,Dong2020,Mok21,Hosseini21}.
\cite{Dong2020} includes a parameter constraint in the supernet training in order to reduce the Lipschitz constant. Also \cite{Hosseini21} adds additional maximization objectives, the certified lower bound and Jacobian norm bound, to the supernet training. \cite{Mok21} includes the loss landscape of the Hessian into the bi-level optimization approach from \cite{2018DARTS}. In contrast to these additional objectives, \cite{Guo2020} proposes adversarial training of the supernet training for increased network robustness.
The recent robustness dataset \cite{Jung2023} facilitates this research area. All $6\,466$ unique architectures in the popular NAS cell-based search space, NAS-Bench-201 \cite{2020NB201}, are evaluated on four different adversarial attacks (FGSM \cite{fgsm}, PGD \cite{pgd}, APGD and Square \cite{croce2020autoattack}) with different attack strengths on three different image datasets, CIFAR-10 \cite{2009CIFAR}, CIFAR-100 \cite{2009CIFAR}, and ImageNet16-120~\cite{2017ImageNet16}. 

\section{Background on Zero-Cost Proxies}\label{sec:ZCP}
As presented in NAS-Bench-Suite-Zero \cite{KrishnakumarWZT22}, we can differentiate the proxies into different types: Jacobian-based ($\color{WildStrawberry}{\bigstar }$), pruning-based ($\color{MidnightBlue}{\blacklozenge}$), baseline ($\color{Mulberry}{\blacksquare}$), piecewise-linear ($\color{Peach}{\blacktriangledown}$), and also Hessian-based ($\color{SeaGreen}{\bullet}$) zero-cost proxies.
In the following, we will provide more information about the ZCPs, which we evaluate in this paper. 
\subsection{Jacobian-based}\label{sec:ZCP_Jacobian}
Mellor et al.~\cite{MellorTSC21} were the first to introduce ZCPs into NAS, by analyzing the network behavior using local linear operations for each input $\textbf{x}_i \in \mathbb{R}^D$ in the training data mini-batch, which can be computed by the Jacobian of the network for each input. The idea is based on the fact that a resulting well-performing untrained network is supposed to be able to distinguish the local linear operations of different data points. For that, the metric \texttt{jacov} was introduced, which is using the correlation matrix of the Jacobian as a Gaussian kernel. The score itself is the Kullback-Leibler divergence between a standard Gaussian and a Gaussian with the mentioned kernel. Therefore, the higher the score, the better the architecture is likely to be.

Building on that, \cite{MellorTSC21} further introduced \texttt{nwot} (Neural Architecture Search without Training), which forms binary codes, depending on whether the rectified linear unit is active or not, which define the linear region in the network. Similar binary codes for two input points indicate that it is more challenging to learn to separate these points. %
\cite{LopesAA21} developed \texttt{nwot} even further by introducing \texttt{epe-nas} (Efficient Performance Estimation Without Training for Neural Architecture Search).
The goal of epe-nas is to evaluate if an untrained network is able to distinguish two points from different classes and equate points from the same class. This is measured by the correlation of the Jacobian (\texttt{jacov}) for input data being from the same class. Therefore, the resulting correlation matrix can be used to analyze how the network behaves for each class and thus may indicate if this network can also distinguish between different classes.

As an alternative to these methods, \cite{AbdelfattahMDL21} proposed a simple proxy, \texttt{grad-norm}, which is simply the sum of the Euclidean norm of the weight gradients. 

So far, these Jacobian-based measurements focused on the correlation with the resulting clean performance of architectures. In this paper, we are also interested in the correlation and influence on the robust accuracy of the architecture. As also used in \cite{Jung2023}, \cite{Hosseini21} combined the search for a high-performing architecture that is also robust against adversarial attacks, by including the Frobenius norm of the Jacobian (\texttt{jacob-fro}) into the search. As introduced in \cite{Hoffman2019} the change of the network output, when a perturbed data point $\textbf{x}_i + \mathbf{\epsilon}, \mathbf{\epsilon} \in \mathbb{R}^D$ is the input to the network instead of the clean data point $\textbf{x}_i$, can be used as a measurement for the robustness of the architecture: the larger the change, the more unstable is the network in case of perturbed input data. This change can be measured by the square of the Frobenius norm of the difference of the network's prediction on perturbed and unperturbed data.

\subsection{Pruning-based}
Pruning-based ZCPs are based on network pruning metrics, which identify the least important parameters in a network at initialization time. 
\cite{LeeAT19} introduced a measurement, \texttt{snip} (Single-shot Network Pruning), which uses a connection sensitivity metric to approximate the change in the loss, when weights with a small gradient magnitude are pruned. 
Based on that, \cite{WangZG20} improves \texttt{snip} by approximating the change in the loss after weight pruning in their \texttt{grasp} (Gradient Signal Preservation) metric. 
 Lastly, \cite{TanakaKYG20} investigated these two pruning-based metrics in terms of layer collapse, and proposes \texttt{synflow}, which multiplies all absolute values of the parameters in the network, which is independent of the training data. 

 In contrast to that, \cite{TurnerCOSG20} obtains and aggregates the Fisher information \texttt{fisher} for all channels in a convolution block,  to identify the channel with the least important effect on the network's loss.

\subsection{Piecewise Linear}
\cite{LinWSCS00021} proposes the \texttt{zen} score, motivated the observation that a CNN can be also seen as a composition of piecewise linear functions being conditioned on activation patterns. Therefore, they propose to measure the network's ability to express complex functions by its Gaussian complexity. This score is data-independent since both the network weights and the input data are sampled from a standard Gaussian distribution.  

\subsection{Hessian-based}\label{sec:ZCP_hessian}
As mentioned in Sec. \ref{sec:ZCP_Jacobian}, the goal for ZCP research was mainly motivated by finding a measurement, which is correlated with the network's performance after training. In this paper, we want to shift the focus towards the robustness of architectures, which is also a crucial aspect of neural architectures. \cite{Mok21} also included the robustness as a target for architecture search, by considering the smoothness of the loss landscape of the architecture. \cite{Zhao2020} investigated the connection between the smoothness of the loss landscape of a network and its robustness, and shows that the adversarial loss is correlated with the biggest eigenvalue of the Hessian. A small Hessian spectrum implies a flat minimum, whereas a large Hessian spectrum, implies a sharp minimum, which is more sensitive to changes in the input and therefore can be more easily fooled by adversarial attacks.

\subsection{Baselines}
In addition to the above mentioned developed ZCPs, basic network information have also been successfully used as ZCPs \cite{AbdelfattahMDL21,NingTLZLYW21}. 
The most common baseline proxies are the number of FLOPS (\texttt{flops}) as well as the number of parameters (\texttt{params}). 
In addition to these, \cite{AbdelfattahMDL21} also considers the sum of the L2-norm of the untrained network weights, \texttt{l2-norm}, and the multiplication of the network weights and its gradients, \texttt{plain}.

\section{Feature Collection and Evaluation}\label{sec:collection}
In the following, we are going to describe our evaluation setting.
\subsection{NAS-Bench-201}
NAS-Bench-201 \cite{2020NB201} is a cell-based search space \cite{White23survey}, in which each cell has 4 nodes and 6 edges. The node represents the architecture's feature map and each edge represents one operation from a predefined operation set. This operation set contains 5 different operations: $1 \times 1~\mathrm{ convolution}, 3 \times 3~\mathrm{ convolution}, 3 \times 3~\mathrm{ avg. pooling}, ~\mathrm{ skip-connection}, ~\mathrm{ zero}$. The cells are integrated into a macro-architecture. The overall search space has a size of $5^6=15\,625$ different architectures, from which $6\,466$ architectures are unique and non-isomorphic. Isomorphic architectures have the same information flow in the network architecture, resulting in similar outcomes and only differ due to numerical reasons. All architectures are trained on three different image datasets, CIFAR-10 \cite{2009CIFAR}, CIFAR-100 \cite{2009CIFAR}, and ImageNet16-120 \cite{2017ImageNet16}. 

\subsection{Neural Architecture Design and Robustness Dataset}
The dataset by \cite{Jung2023} evaluated all the unique architectures in NAS-Bench-201 \cite{2020NB201} against three white-box attacks, i.e., FGSM \cite{fgsm}, PGD \cite{pgd}, APGD \cite{croce2020autoattack}, as well as one black-box attack, Square Attack \cite{croce2020autoattack} to evaluate the adversarial robustness of the architectures, given by the different topologies. In addition, all architectures were also evaluated on corrupted image datasets, CIFAR-10 C and CIFAR-100-C \cite{hendrycks2019robustness}. Along with these evaluations, the dataset also shows three different use cases, on how the data can be used:  evaluation of ZCPs for robustness, NAS on robustness, and an analysis of how the topology and the design of an architecture influence its resulting robustness. This dataset also provides evaluations for the Frobenius norm from Sec. \ref{sec:ZCP_Jacobian} and the biggest eigenvalue of the Hessian from Sec. \ref{sec:ZCP_hessian} as two zero-cost proxies for their first use case. Note, the latter proxy, the Hessian, was only evaluated on the CIFAR-10 image data \cite{2009CIFAR}.

\subsection{Collection}
Both datasets \cite{KrishnakumarWZT22,Jung2023} provide us with the necessary proxies for the NAS-Bench-201 \cite{2020NB201} search space evaluations. Therefore we will analyze 15 different proxies for $6\,466$ architectures on three different image datasets. Note, the ZCPs in \cite{KrishnakumarWZT22} were evaluated on the validation accuracy, while \cite{Jung2023} provide test accuracy information. We will focus here on the test accuracies. 

\section{Evaluations of Zero-Cost Proxies}
As presented in Sec.~\ref{sec:collection}, we can directly gather all proxies. Having these at hand allows us to evaluate the correlation, influence and importance of each proxy not only on the clean test accuracy of the architectures but also on their robust accuracy.

\subsection{Correlation}
\begin{figure}[t!]
\centering
\includegraphics[width=0.7\textwidth]{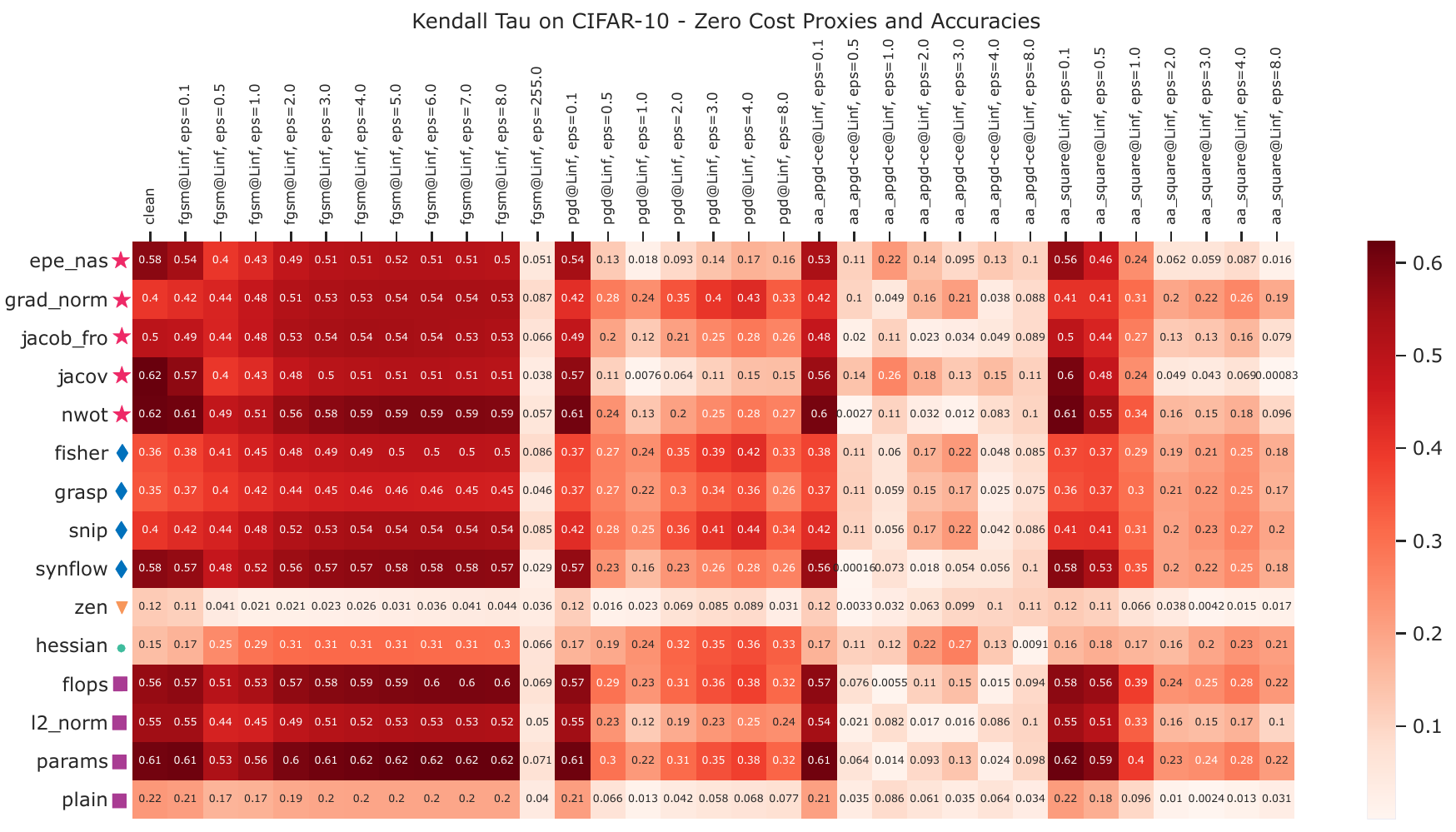}
\caption{Kendall tau rank correlation in \textbf{absolute values} between all zero-cost proxies computed on all architectures given in the robustness dataset \cite{Jung2023} to the test accuracy and adversarial attacks for CIFAR-10.  
\label{fig1:correlation}}
\end{figure}

Figure \ref{fig1:correlation} shows the Kendall tau rank correlation in absolute values between each ZCP and all available accuracies, from clean test accuracy to all adversarial attack test accuracies (\texttt{fgsm}, \texttt{pgd}, \texttt{aa-apgd-ce}, \texttt{aa-square}) on CIFAR-10. For the correlation plots on CIFAR-100 and ImageNet16-120 we refer to the supplementary material. The Jacobian-based proxies (especially \texttt{jacov} and \texttt{nwot}) and the baseline proxies show the highest correlation over all datasets, especially for the FGSM attack.  Furthermore, the large correlation within the same attack over different strengths of $\epsilon$ values stays steady. However, the more difficult the attack gets, from FGSM over PGD to APGD, the larger is the correlations decrease. Interestingly, the \texttt{zen} proxy has the lowest correlation with each test accuracy. 

\begin{table}[t]
    \scriptsize
    \centering
    \caption{Test $R^2$ of the random forest prediction model for both single objective and multi objectives on the clean test accuracy and the robust test accuracy for $\epsilon = 1/255$.}
    \label{tab:R2}
    \resizebox{\columnwidth}{!}{
    \begin{tabular}{l||c|c|c|c|c||c|c|c|c}
    \toprule
        & \multicolumn{9}{c}{\textbf{Test Accuracy $\epsilon= 1/255$}} \\
    \textbf{Dataset} & \multicolumn{5}{c||}{\textbf{Single Objective}} & \multicolumn{4}{c}{\textbf{Multi Objective}} \\
    & \textbf{Clean}  &  \textbf{FGSM} & \textbf{PGD} & \textbf{APGD} & \textbf{Squares} & \textbf{Clean-} & \textbf{Clean-} & \textbf{Clean-} & \textbf{Clean-} \\
    &   &   & &  & & \textbf{FGSM} & \textbf{PGD} & \textbf{APGD} & \textbf{Squares} \\
    \midrule
    CIFAR-10 &  0.97 & 0.88 & 0.65 & 0.66 & 0.75  & 0.92 & 0.81 & 0.82 & 0.86 \\
    CIFAR-100     & 0.96 & 0.75 & 0.69 & 0.73 & 0.68 & 0.86 & 0.83 & 0.85 & 0.82  \\
    ImageNet16-120    &   0.93 & 0.65 & 0.84 & 0.87 & 0.76 & 0.78 & 0.88 & 0.90 & 0.85 \\
    \bottomrule
    \end{tabular}
    }
\end{table}

\begin{figure}[t!]
\centering
\begin{tabular}{@{}c@{}c@{}}
\includegraphics[width=0.5\textwidth]{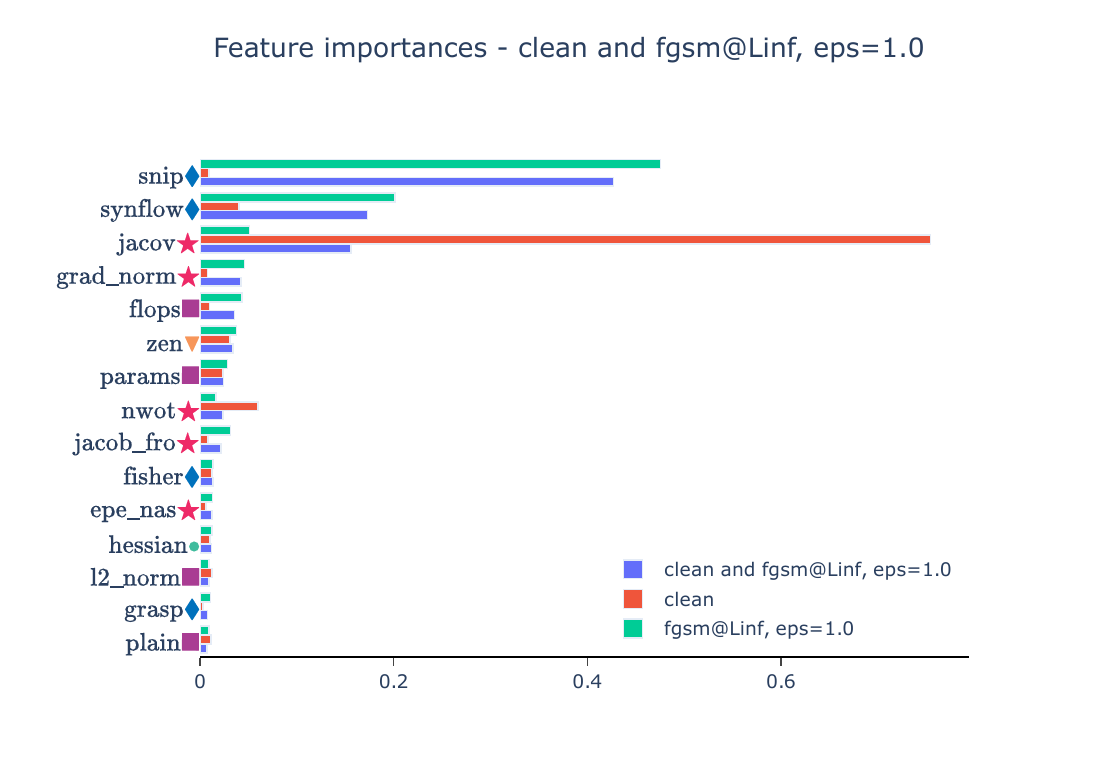}&
\includegraphics[width=0.5\textwidth]{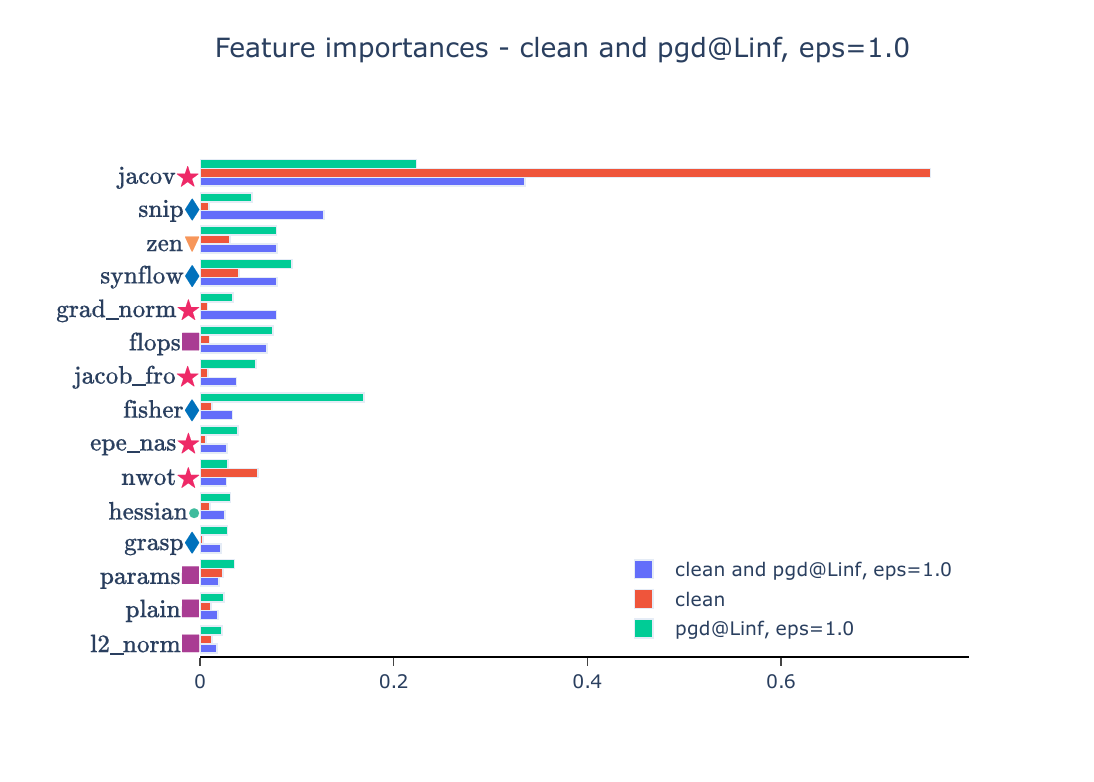} \\
\includegraphics[width=0.5\textwidth]{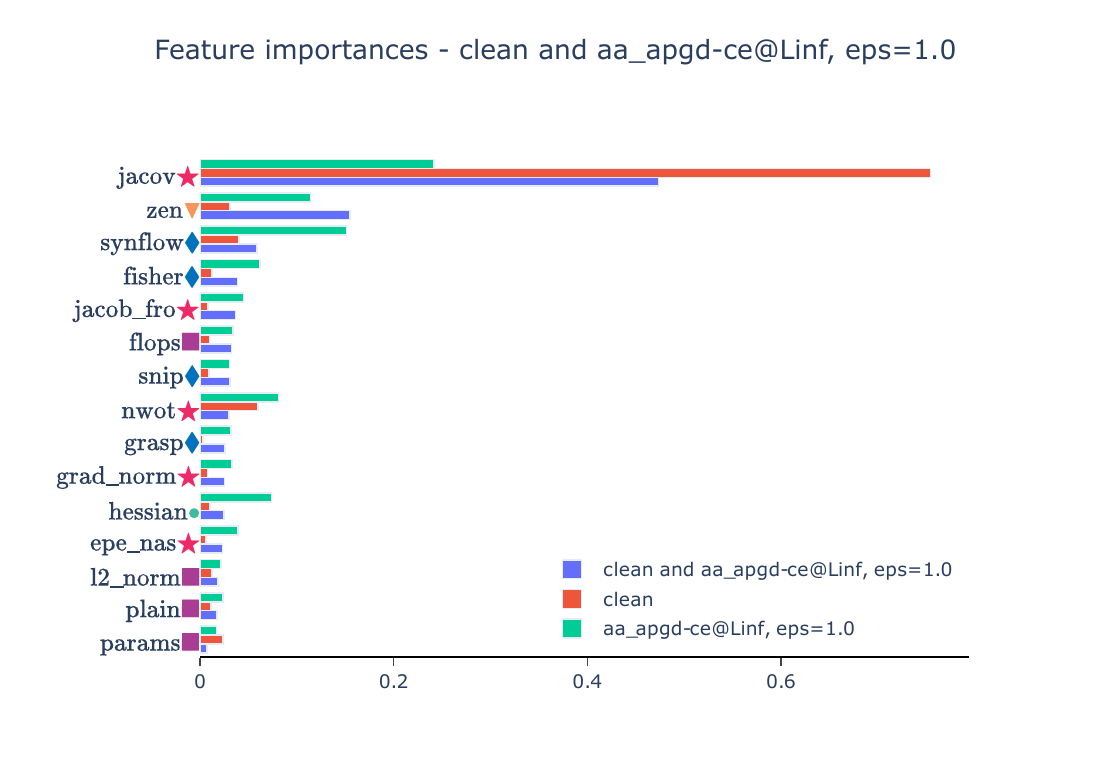}&
\includegraphics[width=0.5\textwidth]{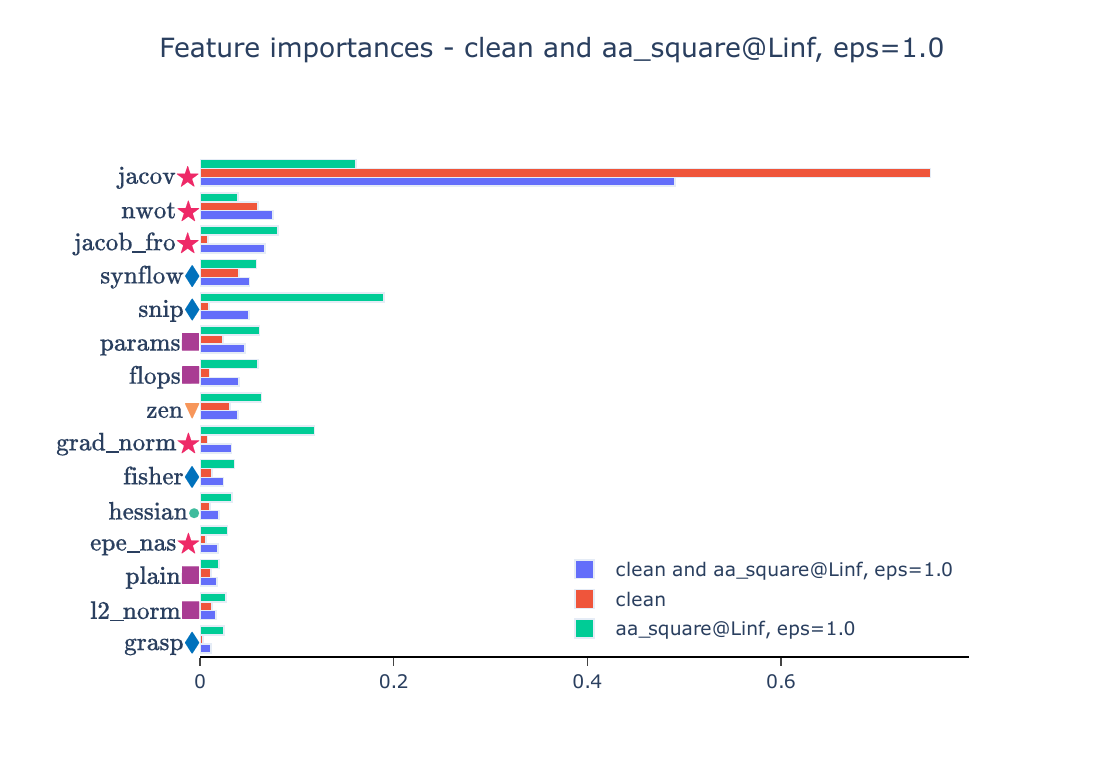}
\end{tabular}
\caption{Feature importance of the random forest prediction model trained on $80\%$ of the data provided in \cite{Jung2023} with all zero-cost proxies as features and multi targets being clean test accuracy and different adversarial accuracies on \textbf{CIFAR-10}.
\label{fig:f_im_cf10}}
\end{figure}

\subsection{Feature Importance}
In the following we want to analyze the ability of ZCPs to be used as features to predict the robust accuracy as a single objective target, as well as the multi-objective target, for both clean and robust accuracy. 

For that we split the architecture dataset ($6\,466$ architectures from \cite{Jung2023}) into $80\%$ training and $20\%$ test data to train a random forest regression model, that takes as input a feature vector of all concatenated ZCPs. We use the default regression parameters, with $100$ numbers of trees and the mean squared error as the target criterion. For a better overview, we will consider the attack $\epsilon$ value of $1/255$ for all the following experiments. 
We first analyze the prediction ability itself in Tab.~\ref{tab:R2} by means of the $R^2$ score of the prediction model on the test dataset. As we can see, the random forest prediction model is able to predict the single accuracy (clean and robust) and the multi-objective accuracies in a proper way, while the prediction of the clean accuracy seems to be the easiest task, where $R^2$ values of 0.93 to 0.97 are reached.

Next, we are interested in the feature importance of individual ZCPs. So far the ZCPs in NAS-Bench-Suite-Zero are motivated by their correlation with the resulting clean accuracy of the architectures. Yet, how well can they be transferred to the more challenging task of robust accuracy and even more challenging than that, the multi-objective task?
High clean accuracy does not necessarily mean that a network is also robust. Therefore we use all ZCPs from NAS-Bench-Suite-Zero \cite{KrishnakumarWZT22} and the robustness dataset \cite{Jung2023}, which we already presented in Sec.~\ref{sec:ZCP} as feature inputs for the random forest prediction model and calculate their importance as the mean variance reduction (Gini importance). Note that in this measure, the comparison of correlated features needs to be considered with caution.
Fig.~\ref{fig:f_im_cf10} visualizes the feature importance on the CIFAR-10 image datasets for all adversarial attacks with $\epsilon=1/255$ for all three use cases: clean test accuracy, robust test accuracy, both test and robust accuracy. The latter one is used as the bar plot alignment. As we can see, \texttt{jacov} is the most important feature for both the clean test accuracy and the multi-objective task, except for FGSM. Interestingly, the Top 5 most important ZCPs are all different for all considered adversarial attacks.

\begin{figure}[t!]
\centering
\begin{tabular}{@{}c@{}c@{}}
\includegraphics[width=0.5\textwidth]{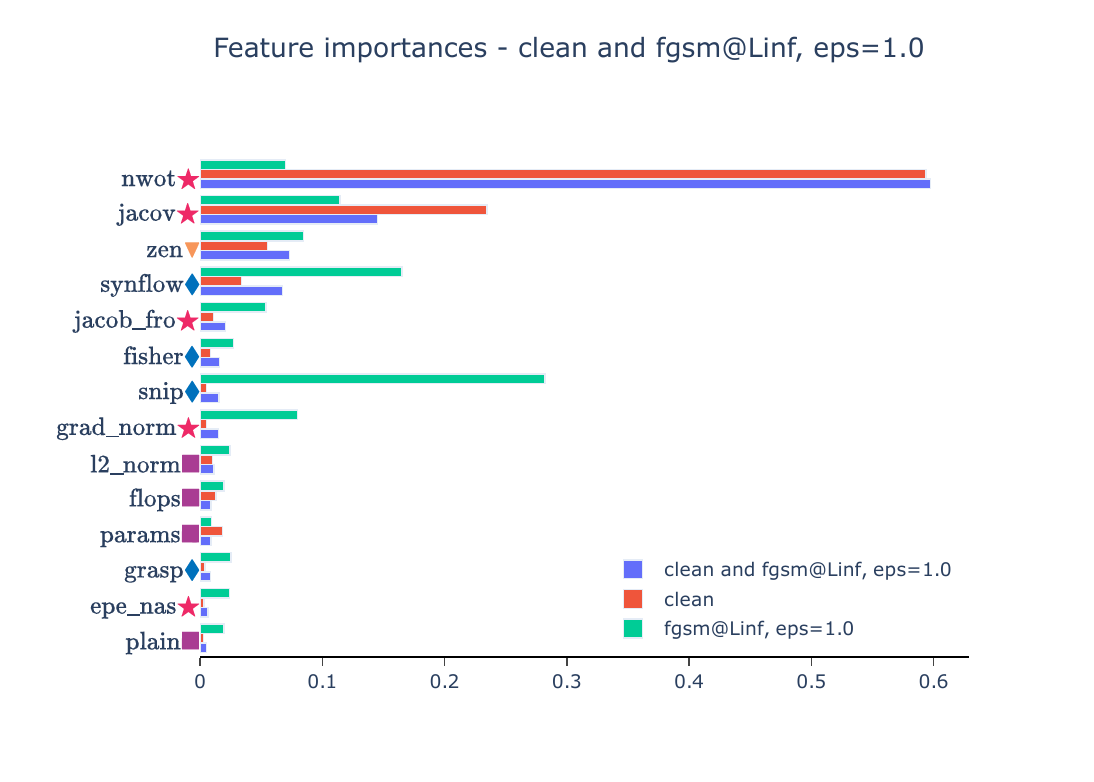}&
\includegraphics[width=0.5\textwidth]{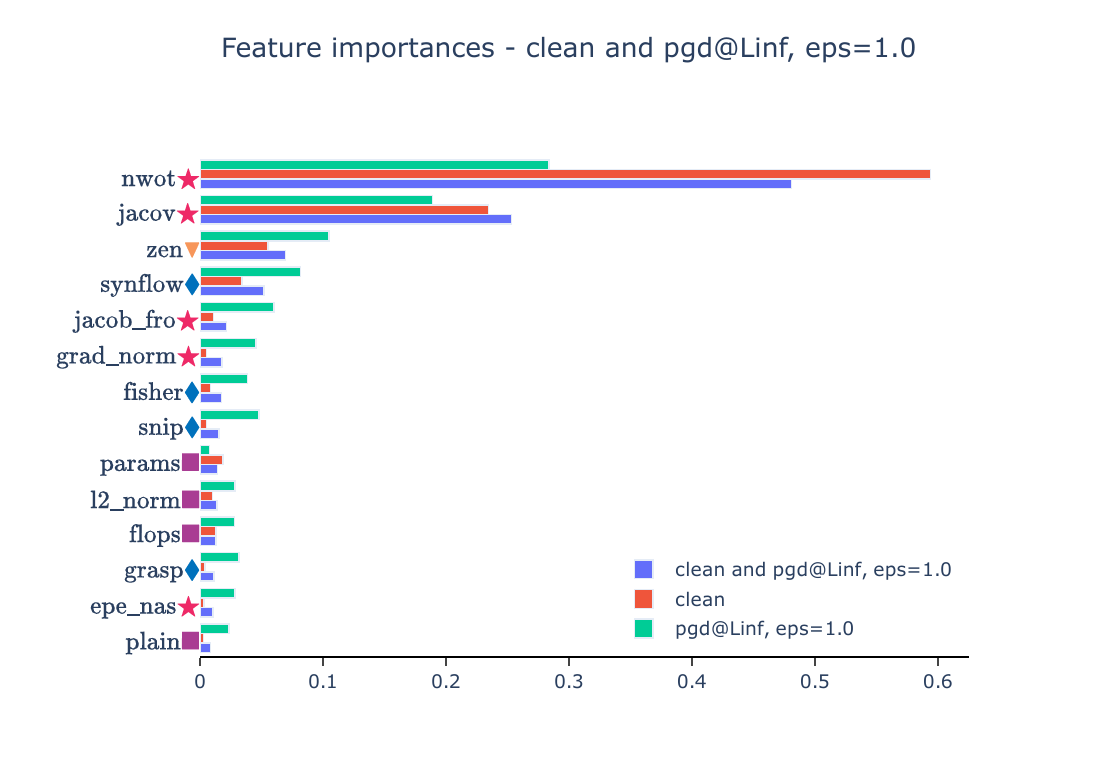} \\
\includegraphics[width=0.5\textwidth]{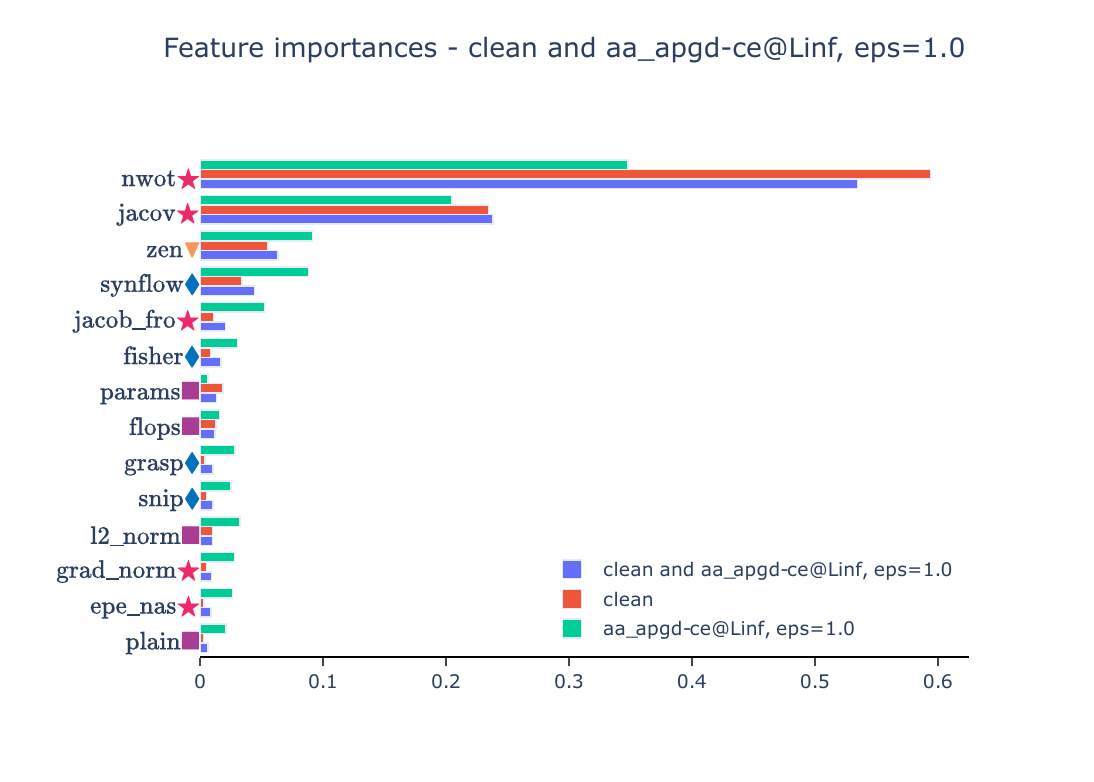}&
\includegraphics[width=0.5\textwidth]{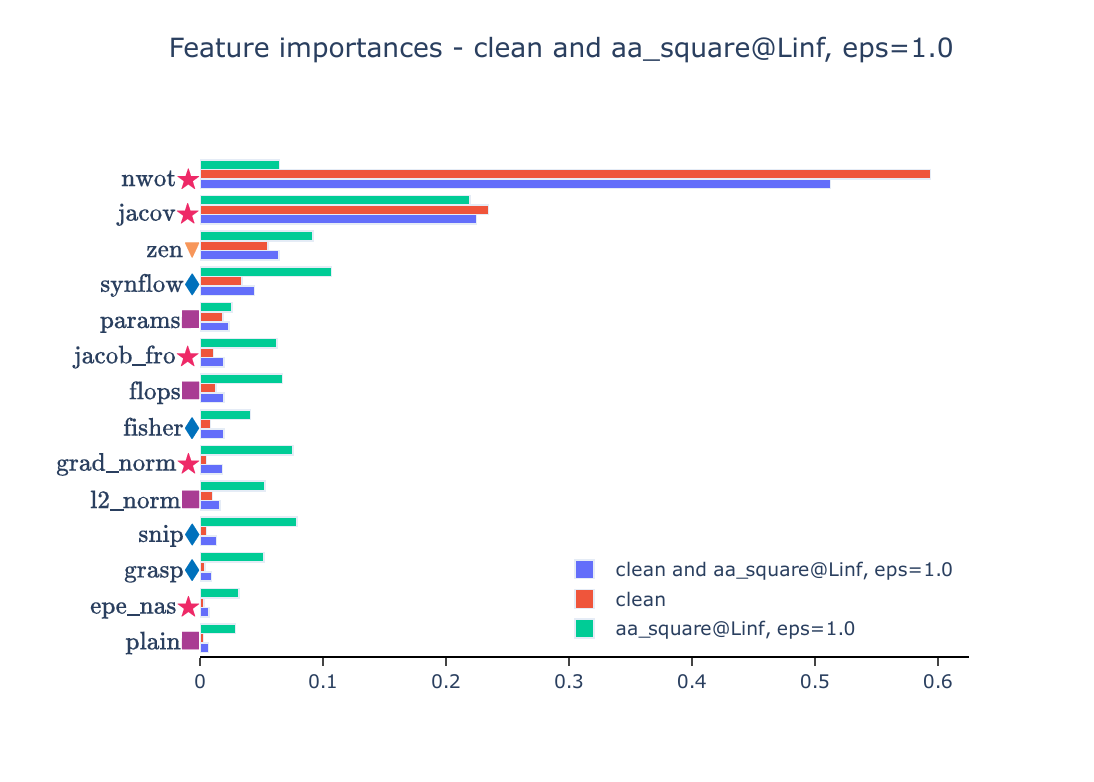}
\end{tabular}
\caption{Feature importance of the random forest prediction model trained on $80\%$ of the data provided in \cite{Jung2023} with all zero-cost proxies as features and multi targets being clean test accuracy and different adversarial accuracies on \textbf{CIFAR-100}.   
\label{fig:f_im_cf100}}
\end{figure}

If we look at the feature importance for CIFAR-100 (Fig.~\ref{fig:f_im_cf100}), we can see that the Top 4 most important features for clean and robust accuracy are always the same, with the Jacobian-based ZCP \texttt{nwot} being the most important one. 
Note here, for CIFAR-100 the rank correlation decreases the most for more difficult attacks.  

\begin{figure}[t!]
\centering
\begin{tabular}{@{}c@{}c@{}}
\includegraphics[width=0.5\textwidth]{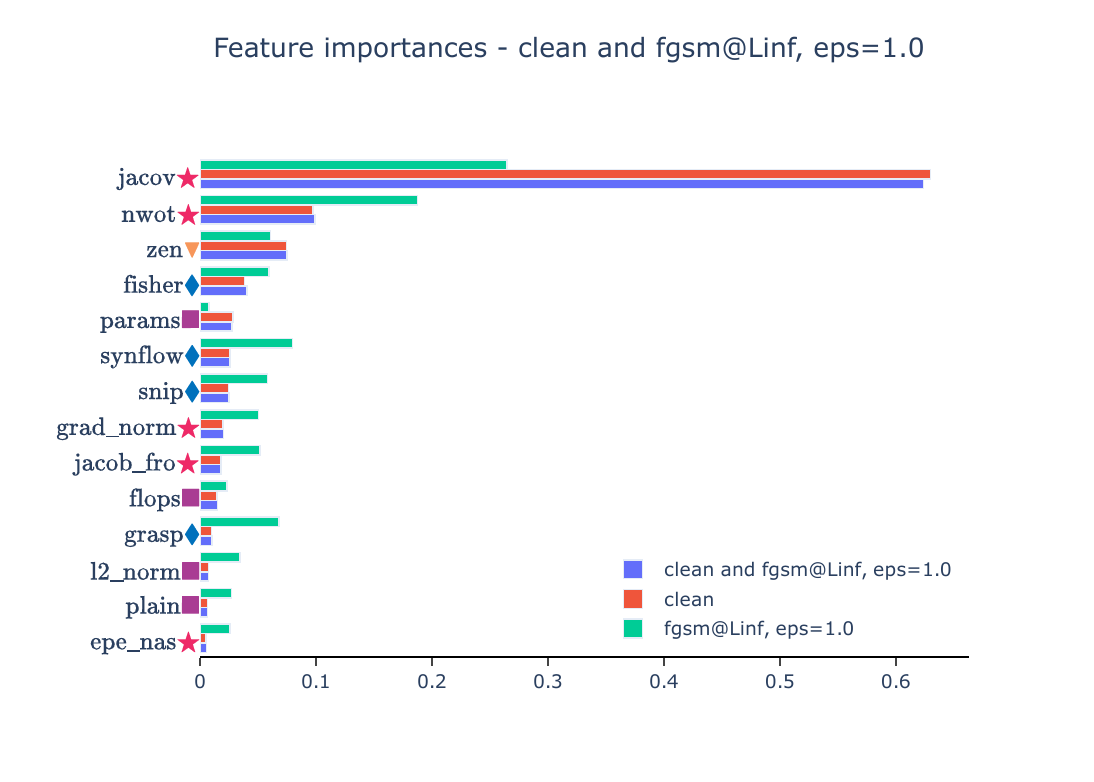}&
\includegraphics[width=0.5\textwidth]{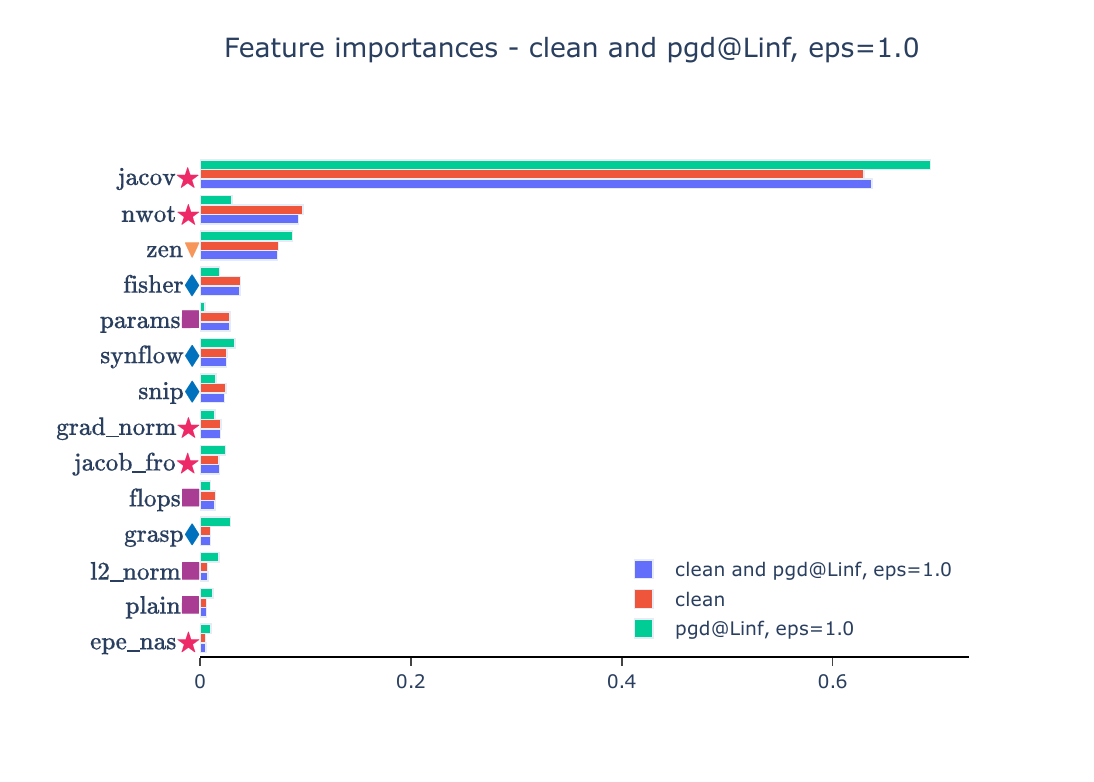}\\
\includegraphics[width=0.5\textwidth]{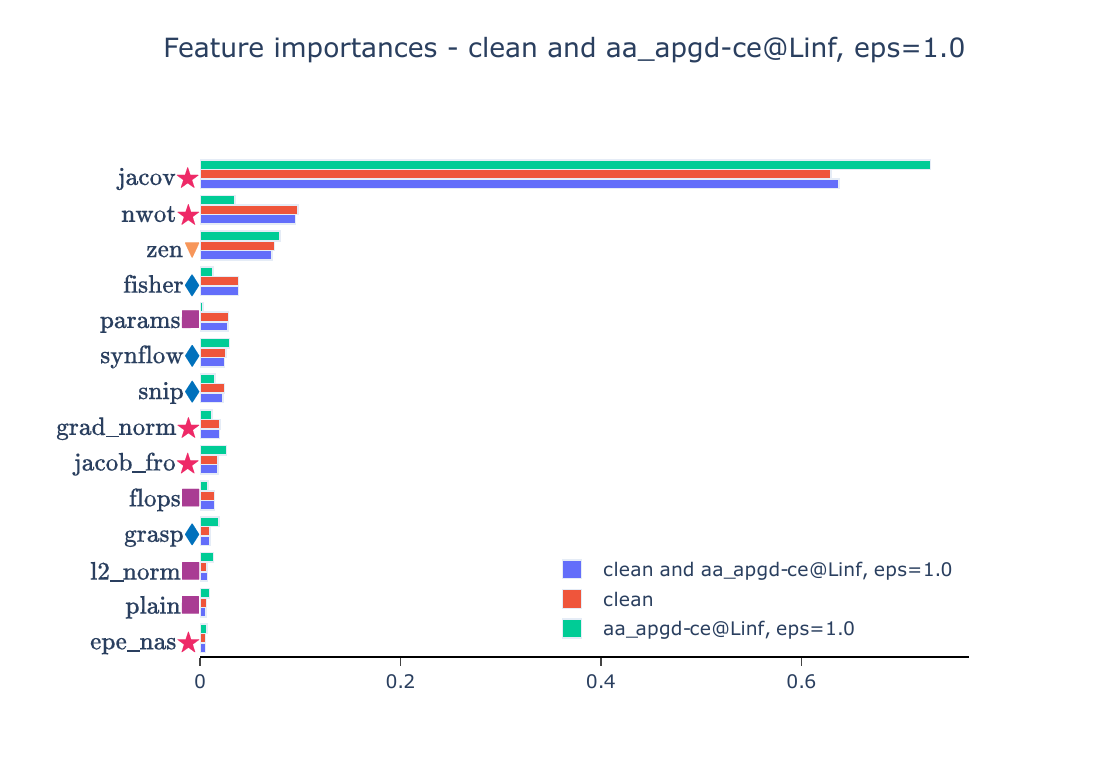}&
\includegraphics[width=0.5\textwidth]{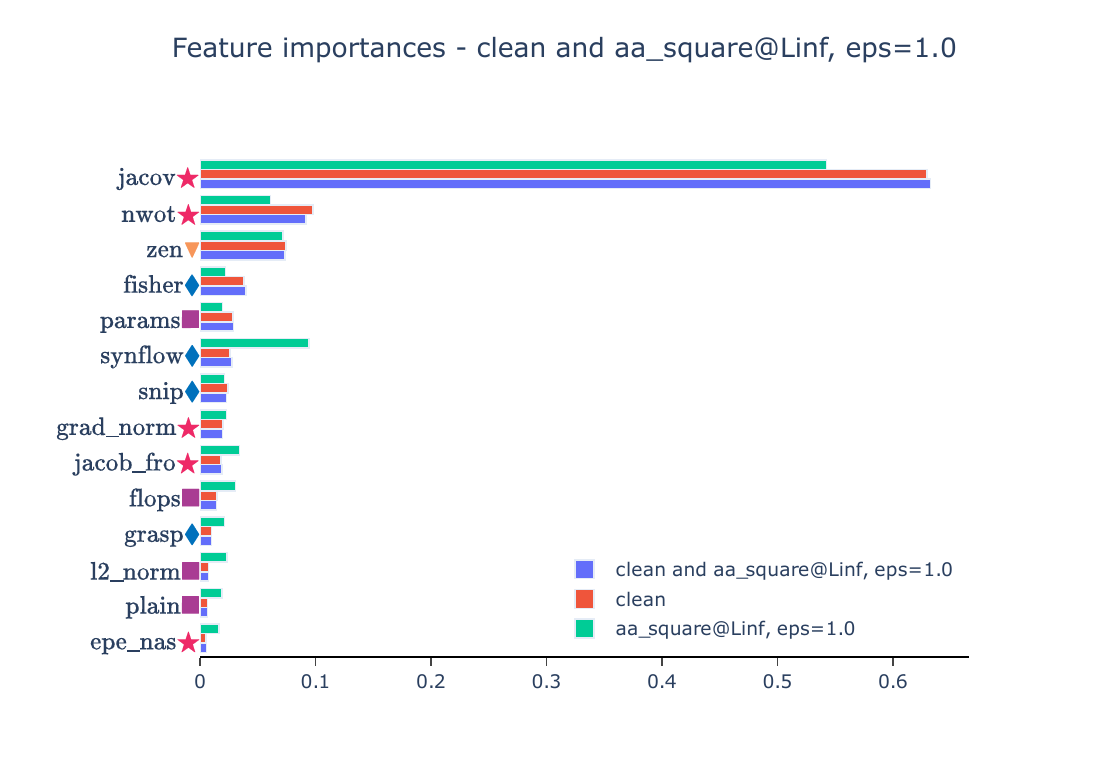}
\end{tabular}
\caption{Feature importance of the random forest prediction model trained on $80\%$ of the data provided in \cite{Jung2023} with all zero-cost proxies as features and multi targets being clean test accuracy and different adversarial accuracies on \textbf{ImageNet16-120}.   
\label{fig:f_im_im16}}
\end{figure}

\begin{table}[t]
    \scriptsize
    \centering
    \caption{Test $R^2$ of the random forest prediction model using the \textbf{most important feature} for both single objectives and multi objectives on the clean test accuracy and the robust test accuracy for $\epsilon = 1/255$.}
    \label{tab:R2_one_feature}
    \resizebox{\columnwidth}{!}{
    \begin{tabular}{l||r|r|r|r|r||r|r|r|r}
    \toprule
        & \multicolumn{9}{c}{\textbf{Test Accuracy $\epsilon= 1/255$}} \\
    \textbf{Dataset} & \multicolumn{5}{c||}{\textbf{Single Objective}} & \multicolumn{4}{c}{\textbf{Multi Objective}} \\
    & \textbf{Clean}  &  \textbf{FGSM} & \textbf{PGD} & \textbf{APGD} & \textbf{Squares} & \textbf{Clean-} & \textbf{Clean-} & \textbf{Clean-} & \textbf{Clean-} \\
    &   &   & &  & & \textbf{FGSM} & \textbf{PGD} & \textbf{APGD} & \textbf{Squares} \\
    \midrule
    CIFAR-10 &  {\color{WildStrawberry}{\texttt{jacov}}} 0.84 & {\color{MidnightBlue}{\texttt{snip}}} 0.41 & {\color{WildStrawberry}{\texttt{jacov}}} -0.31 & {\color{WildStrawberry}{\texttt{jacov}}} 0.002 & {\color{MidnightBlue}{\texttt{snip}}} 0.16  &  {\color{MidnightBlue}{\texttt{snip}}} 0.37 & {\color{WildStrawberry}{\texttt{jacov}}}0.26 & {\color{WildStrawberry}{\texttt{jacov}}} 0.42 & {\color{WildStrawberry}{\texttt{jacov}}} 0.54 \\
    CIFAR-100     & {\color{WildStrawberry}{\texttt{nwot}}} 0.74 & {\color{MidnightBlue}{\texttt{snip}}} 0.12 & {\color{WildStrawberry}{\texttt{nwot}}} 0.09 & {\color{WildStrawberry}{\texttt{nwot}}} 0.28 & {\color{WildStrawberry}{\texttt{jacov}}} -0.004 & {\color{WildStrawberry}{\texttt{nwot}}} 0.42 & {\color{WildStrawberry}{\texttt{nwot}}} 0.42 & {\color{WildStrawberry}{\texttt{nwot}}} 0.51 &{\color{WildStrawberry}{\texttt{nwot}}} 0.31  \\
    ImageNet16-120    &{\color{WildStrawberry}{\texttt{jacov}}}   0.61 &{\color{WildStrawberry}{\texttt{jacov}}} -0.003 &{\color{WildStrawberry}{\texttt{jacov}}} 0.58 &{\color{WildStrawberry}{\texttt{jacov}}} 0.64 &{\color{WildStrawberry}{\texttt{jacov}}} 0.44 &{\color{WildStrawberry}{\texttt{jacov}}} 0.31 &{\color{WildStrawberry}{\texttt{jacov}}} 0.60 &{\color{WildStrawberry}{\texttt{jacov}}} 0.63 &{\color{WildStrawberry}{\texttt{jacov}}} 0.52 \\
    \bottomrule
    \end{tabular}
    }
\end{table}

When we turn now to ImageNet16-120, the ranking of the feature importance is the same for all considered attacks. Also there, a Jacobian-based ZCP, in this case \texttt{jacov}, is the most important feature. 

In total, the same features being the most important for the clean accuracy are also important for the architecture's robust accuracy. 

Interestingly, we can see that the task of robust accuracy prediction becomes more difficult on all three image datasets and attacks (see smaller $R^2$ values in Tab. \ref{tab:R2}). 
In addition, Fig. \ref{fig:f_im_cf10}, \ref{fig:f_im_cf100}, and \ref{fig:f_im_im16} give the impression that solely the most important feature could be used to predict the clean accuracy, but not the robust accuracy.

As we can see in Tab. \ref{tab:R2_one_feature}, the joint consideration of several proxies is necessary in terms of $R^2$ to predict a model's robustness, while the clean accuracy can be regressed from only one such feature.

\subsection{Feature Importance excluding Top 1}
The previous section results in the fact that only one proxies is not sufficient to predict the robustness of an architecture. This leads to the next question if the regression model can still make good predictions, using all ZCPs \textbf{but} the most important one.

\begin{table}[ht!
]
    \scriptsize
    \centering
    \caption{Test $R^2$ of the random forest prediction model for the multi-objective task for $\epsilon = 1/255$ \textbf{without the most important feature} for the FGSM and PGD, respectively.}
    \label{tab:R2_wo_top1}
    \begin{tabular}{l||c|c}
    \toprule
        & \multicolumn{2}{c}{\textbf{Test Accuracy $\epsilon= 1/255$}} \\
    \textbf{Dataset} &  \multicolumn{2}{c}{\textbf{Multi Objective}} \\
    &\textbf{Clean-FGSM} & \textbf{Clean-PGD} \\
    \midrule
    CIFAR-10 & 0.92 & 0.79   \\
    CIFAR-100     & 0.87 & 0.83 \\
    ImageNet16-120    &0.72 & 0.81 \\
    \bottomrule
    \end{tabular}
\end{table}

The comparison of Tab. \ref{tab:R2} with Tab.\ref{tab:R2_wo_top1} presents only a slight decrease in $R^2$ values.
The prediction ability for the clean accuracy and FGSM robust accuracy on CIFAR-10, as well as for the clean accuracy and PGD accuracy on CIFAR-100 even remained the same.

\begin{figure}[ht!]
\centering
\includegraphics[width=0.45\textwidth]{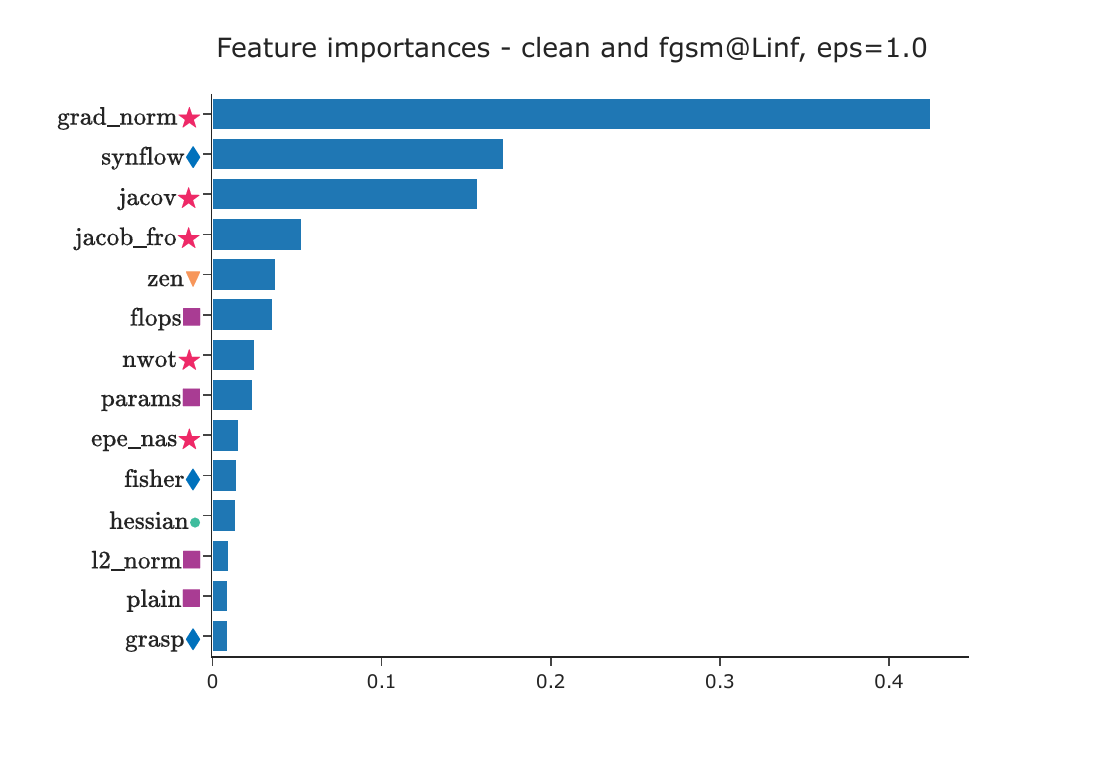}
\includegraphics[width=0.45\textwidth]{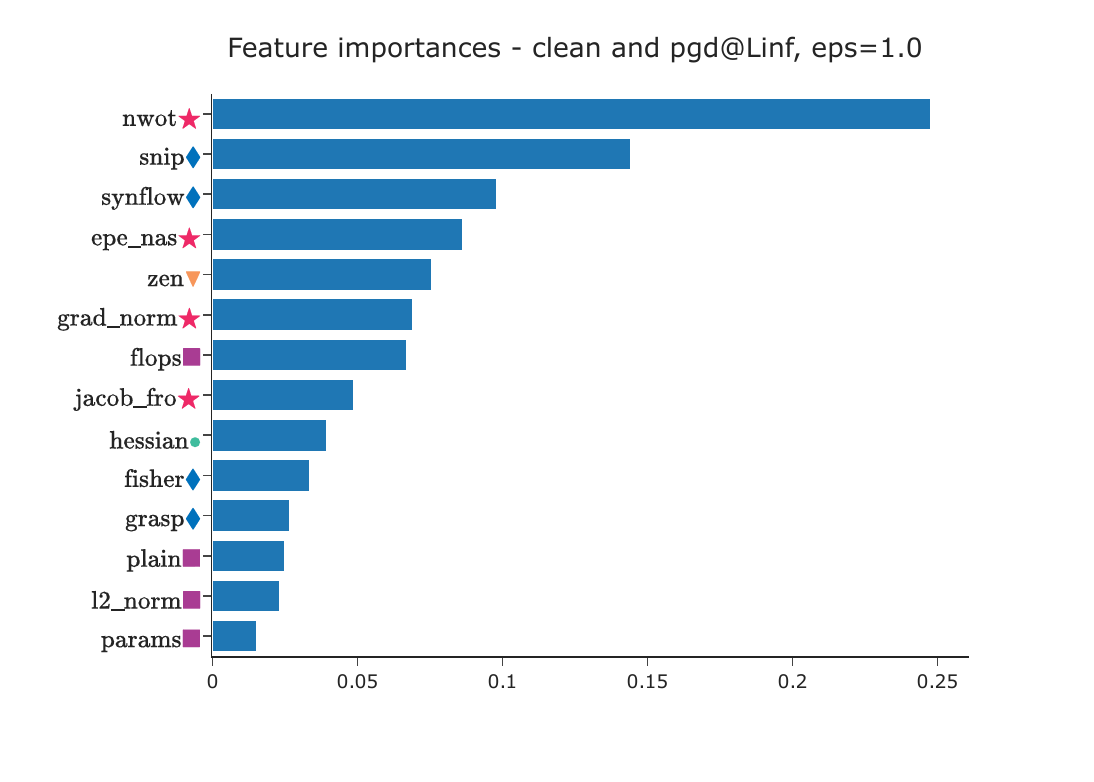}
\includegraphics[width=0.45\textwidth]{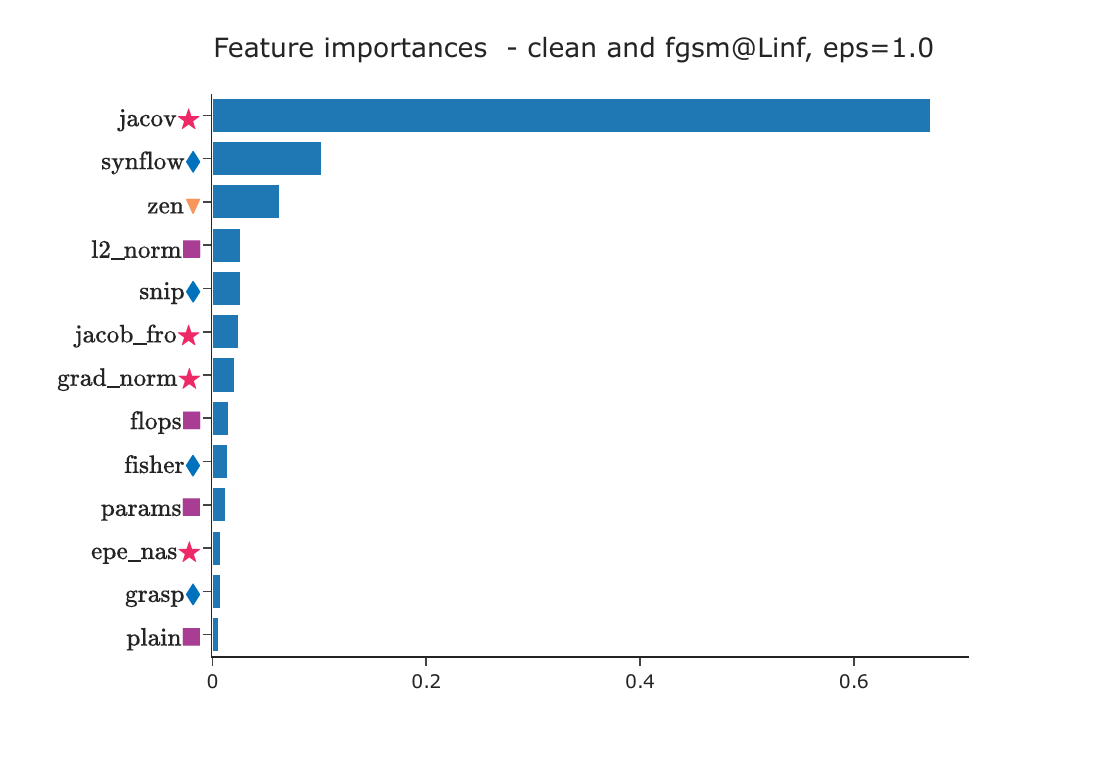}
\includegraphics[width=0.45\textwidth]{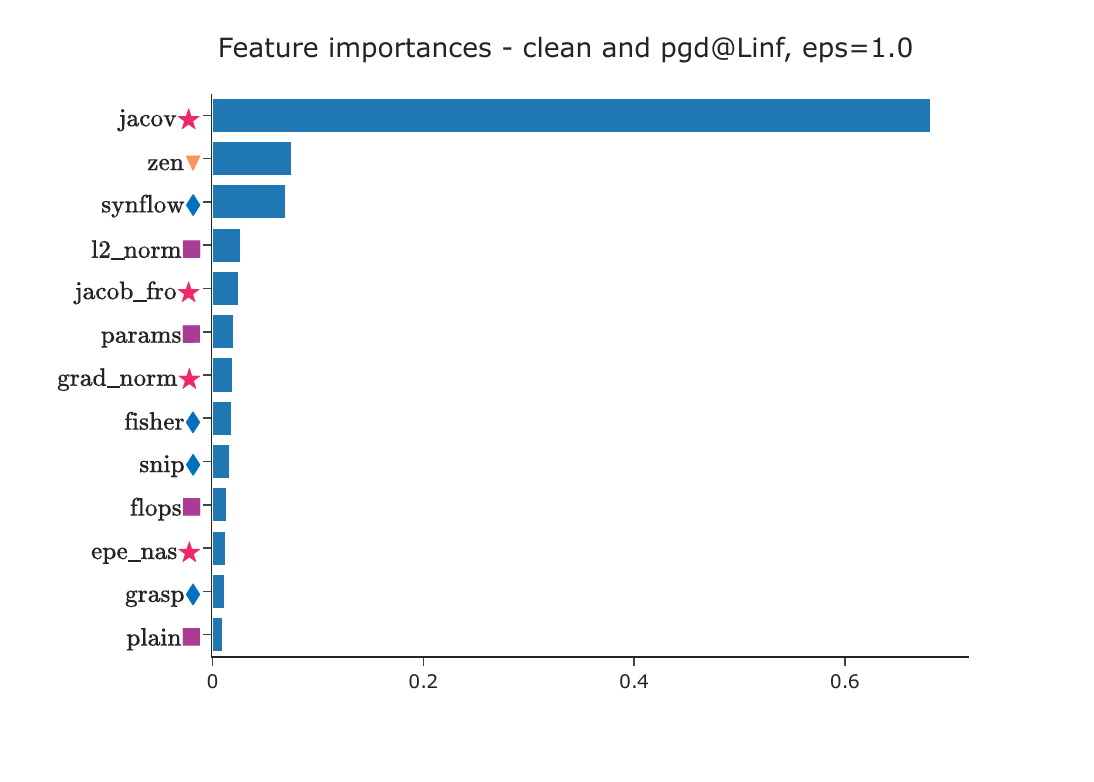}
\includegraphics[width=0.45\textwidth]{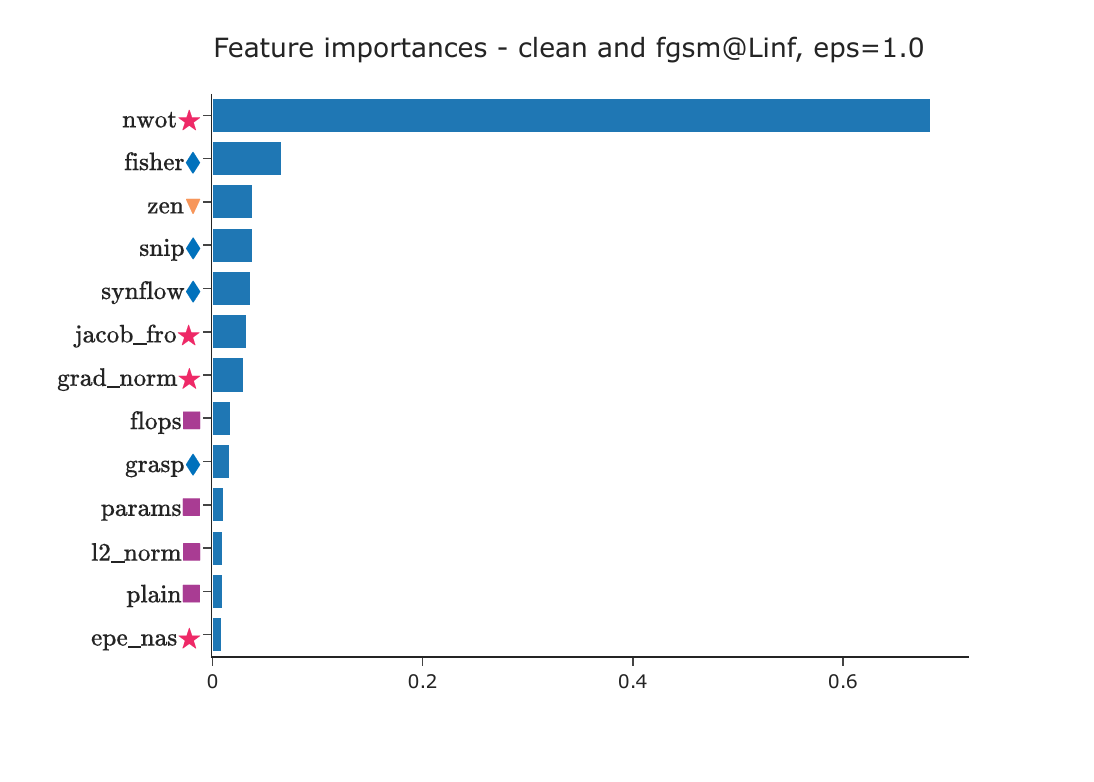}
\includegraphics[width=0.45\textwidth]{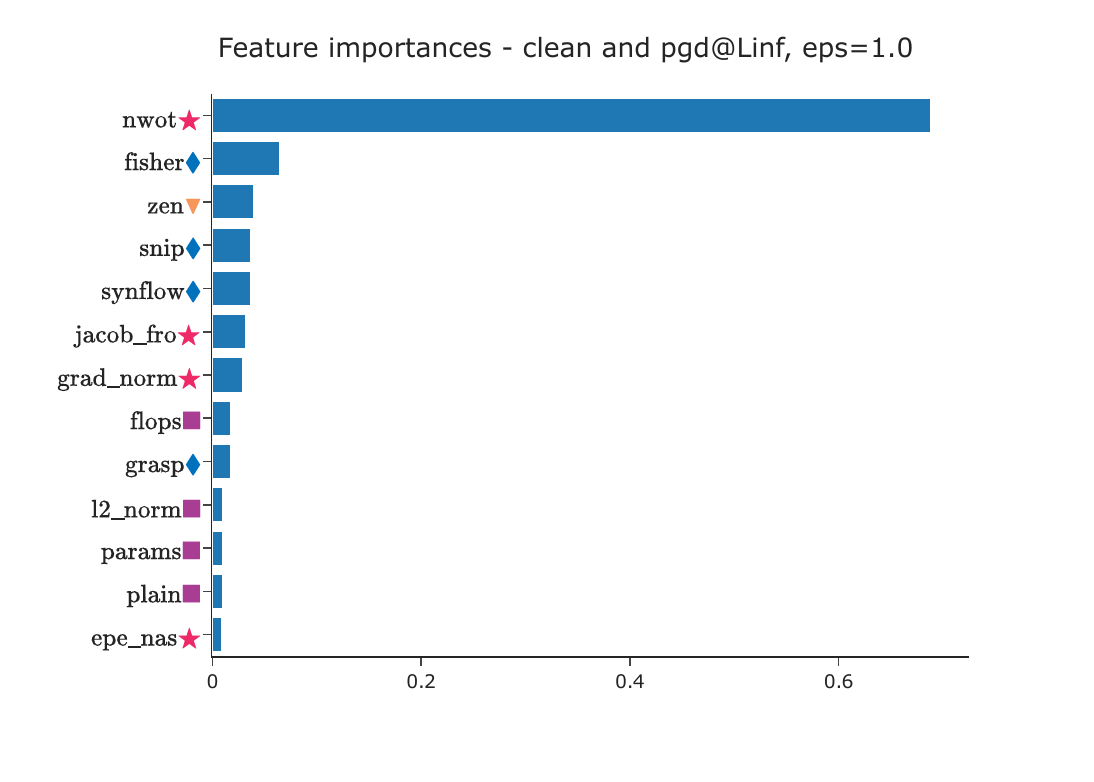}
\caption{Feature importance of the random forest prediction model trained on $80\%$ of the data provided in \cite{Jung2023} \textbf{without the most important feature} from Tables \ref{fig:f_im_cf10}, \ref{fig:f_im_cf100}, \ref{fig:f_im_im16} for FGSM, and PGD at $\epsilon=1/255$. (\textbf{top}): CIFAR-10, (\textbf{middle}): CIFAR-100, (\textbf{bottom}): ImageNet16-120.  
\label{fig1:f_im_2}}
\end{figure}

We also plot the updated feature importances in Fig. \ref{fig1:f_im_2} for all ZCPs without the most important one from the previous section for FGSM and PGD. 
CIFAR-100 (middle) and ImageNet16-120 (bottom) show a new most distinctive important feature, which is in both cases a Jacobian-based one.
If we turn towards CIFAR-10 (top) there is no such distinctive important feature, and rather a steady decrease in importance from one feature to the other.

\section{Conclusion}
In this paper, we presented an analysis of 15 zero-cost proxies with regards to their ability to act as performance prediction techniques for architecture robustness using a random forest. We made use of two provided datasets, NAS-Bench-Suite-Zero and a robustness dataset, which allows fast evaluations on the here-considered NAS-Bench-201 search space. Commonly, these zero-cost proxies are targeting the clean accuracy of architectures. Therefore, our analysis shows that the prediction of robustness is a more difficult task. Additionally, we investigated the feature importance of these zero-cost proxies, which led to the finding, that only one feature is not sufficient to predict the robustness, and that the regression model tends to rely on all available features. 

\subsection*{Acknowledgment}
The authors acknowledge support by the DFG research unit 5336 Learning to Sense.
\FloatBarrier

%
%
%
\bibliographystyle{splncs04}
\bibliography{egbib}

\newpage
\appendix
\section{Background on Robustness Dataset}
In this section, we will provide more information about the adversarial attacks from \cite{Jung2023}.

\subsection{FGSM}
\cite{fgsm} proposed a "fast gradient sign method'' to generate adversarial examples $\tilde{\mathbf{x}}$ with:
\begin{equation}
    \tilde{\mathbf{x}} = \mathbf{x} + \epsilon ~\mathrm{sign}(\nabla_\mathbf{x} J(\mathbf{\theta},\mathbf{x},y)),
\end{equation}
where $\mathbf{x}$ is the clean input image and $y$ the corresponding target for $\mathbf{x}$, $\theta$ are the network parameters and $\epsilon$ the perturbation magnitude. $J(\mathbf{\theta},\mathbf{x},y)$ is the cost function to train the network, which is cross-entropy in \cite{Jung2023}. 
The robustness dataset considers different perturbations magnitudes $\{0.1,0.5,1,2,3,\ldots,8\}/255$ for all architectures and image datasets.

\subsection{PGD}
In contrast, PGD \cite{pgd} perturbs the image in an iterative manner with small step size $\alpha$ and clips the results after each step:
\begin{equation}
    \mathbf{x}_0 =\mathbf{x}, \quad \tilde{\mathbf{x}}_{n+1} = \mathrm{clip}_{\mathbf{x},\epsilon}  (\tilde{\mathbf{x}}_n + \alpha \mathrm{sign}(\nabla_{\mathbf{x}} J(\mathbf{\theta},\tilde{\mathbf{x}}_n,y)))
\end{equation}
where $\mathrm{clip}_{\mathbf{x},\epsilon}(\cdot)$ clips the function to be in an $L_{\infty}$-$\epsilon$ neighborhood of the input image $\mathbf{x}$.  \cite{Jung2023} considers different $\epsilon$ magnitudes $\{0.1,0.5,1,2,3,4,8\}/255$ for all architectures and image datasets. The step size $\alpha$ is set to $0.01/0.3$ and the amount of iterations is set to $40$.

\subsection{APGD}
\cite{croce2020autoattack} provides PGD in an adaptive manner (APGD) by also reducing the step size in an iterative manner.  \cite{Jung2023} uses $100$ as the number of attack iterations and the same $\epsilon$ values as for PGD. 

\subsection{Square}
Square \cite{croce2020autoattack} is a black-box attacks, which does not have access to the gradient information of the network. For this attack, the goal is to change the correctly predicted class for the input point x, by solvig the following optimization problem:
\begin{equation}
    \underset{\tilde{x}}{\mathrm{min}} \{f_y(\tilde{x}) - \underset{k \neq y}{\mathrm{max}} f_k(\tilde{x})\}, \quad  \mathrm{s.t.}  \quad \Vert \tilde{x}-x \Vert_p  \leq \epsilon,
\end{equation}
where $p$ is the $L_p$- norm bound, which is in \cite{Jung2023} set at the most commonly used $L_\infty$ norm, $f_k(x)$ are the predictions of the network for class $k$ given the input $x$.
Also here the $\epsilon$ values are the same as for PGD, with $5\,000$ search iterations.  

\section{Correlation}
We provide here the additional correlation plots for CIFAR-100 and ImageNet16-120.
\begin{figure}[ht!]
\centering
\includegraphics[width=0.8\textwidth]{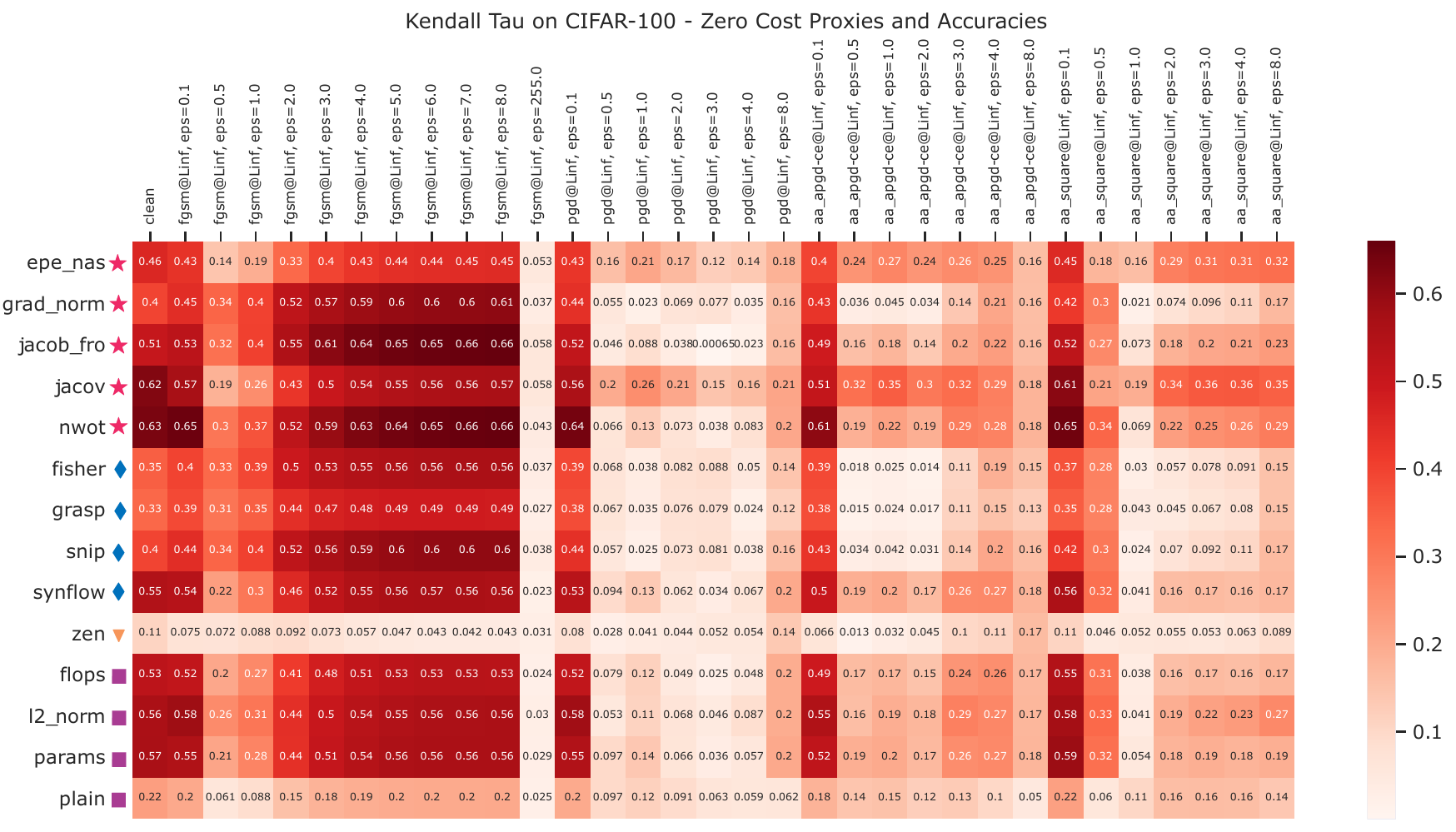}
\includegraphics[width=0.8\textwidth]{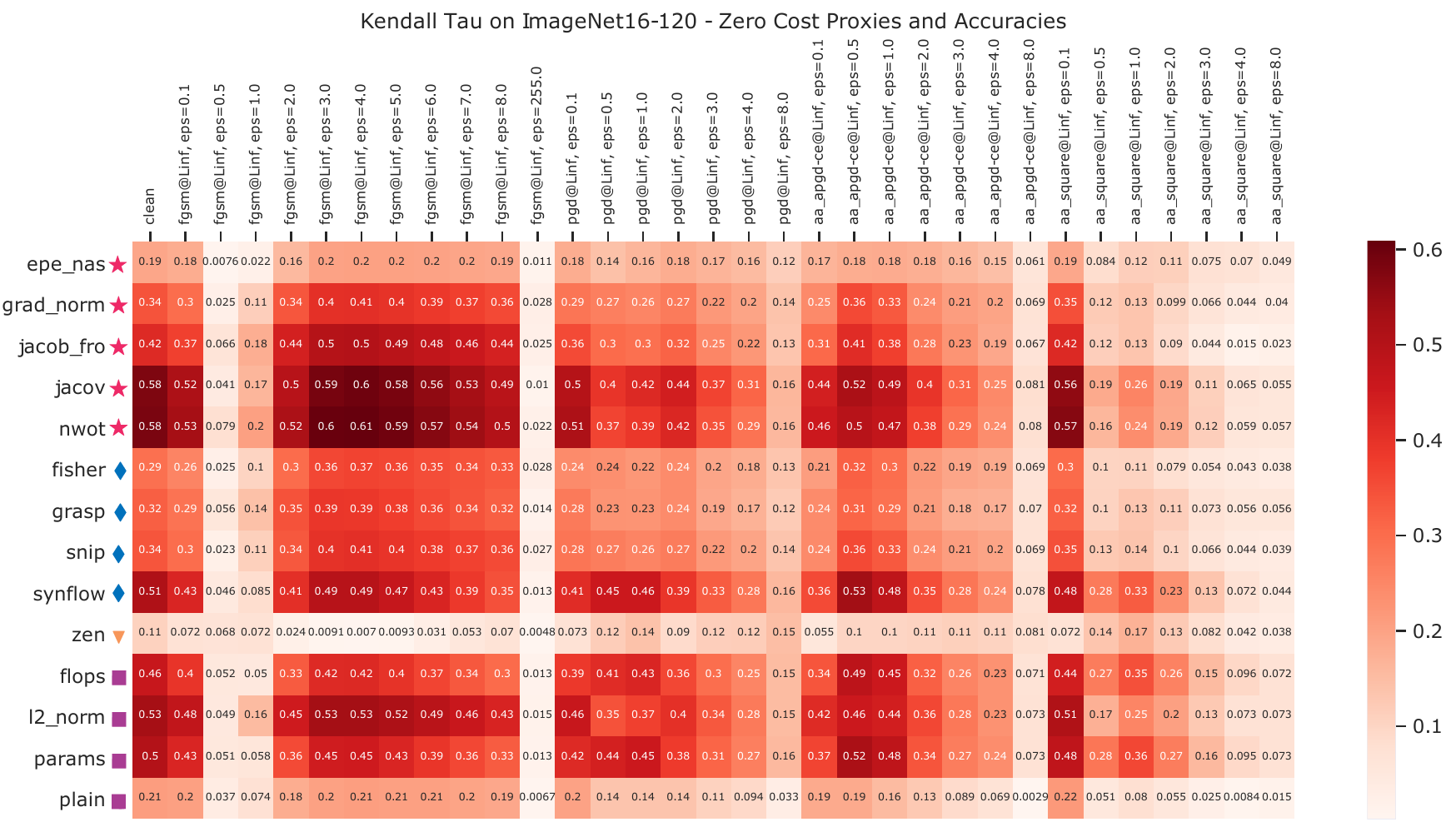}
\caption{Kendall tau rank correlation in \textbf{absolute values} between all zero-cost proxies computed on all architectures given in the robustness dataset \cite{Jung2023} to the test accuracy and adversarial attacks for (\textbf{top}) CIFAR-100, and (\textbf{bottom}) ImageNet16-120.  
\label{fig1:correlation}}
\end{figure}

\section{Permutation Importance}
\begin{figure}[ht!]
\centering
\includegraphics[width=0.45\textwidth]{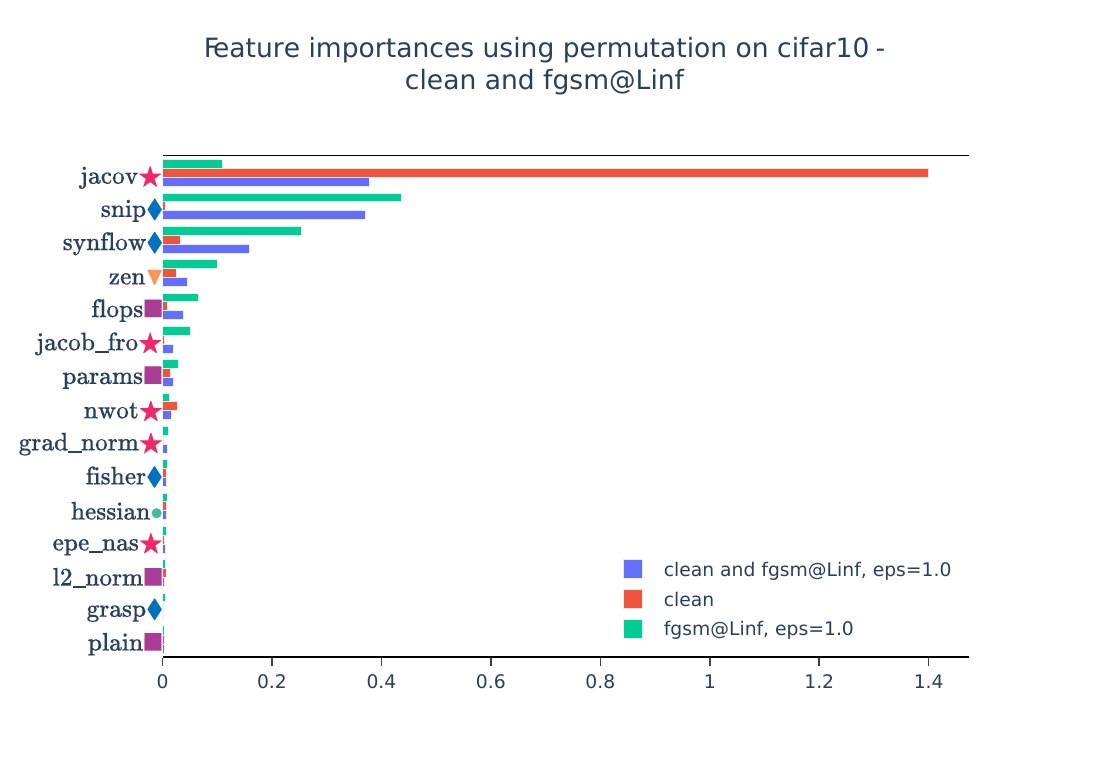}
\includegraphics[width=0.45\textwidth]{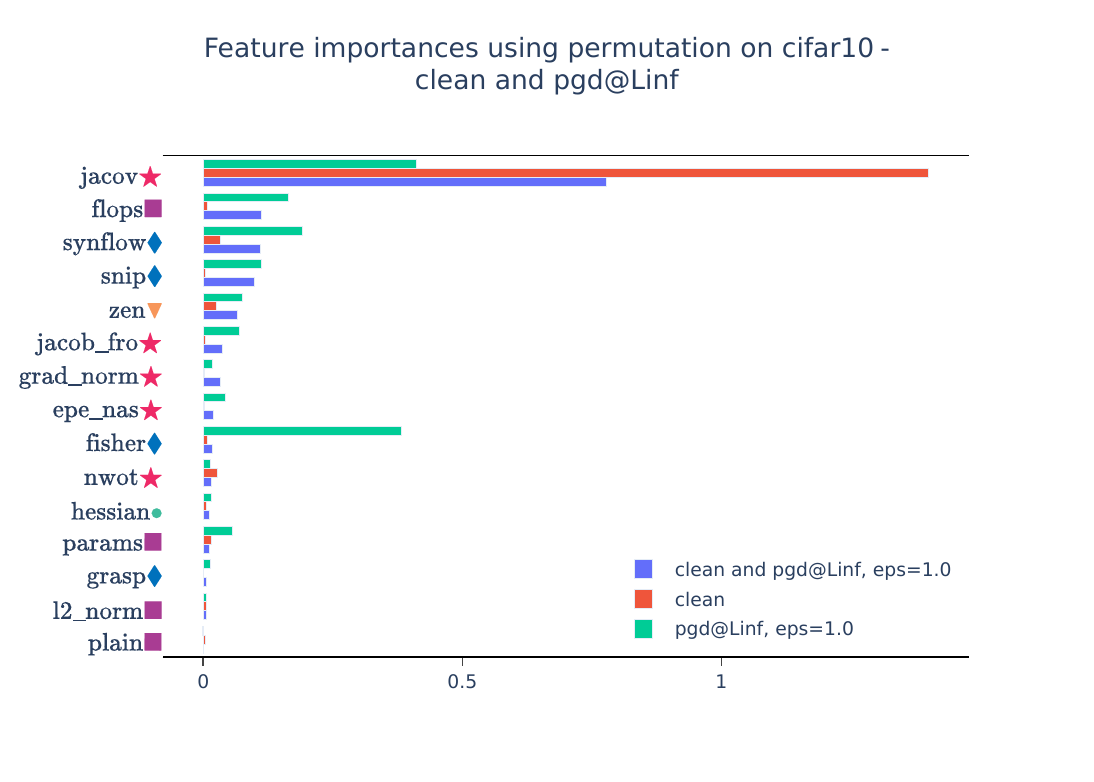}
\includegraphics[width=0.45\textwidth]{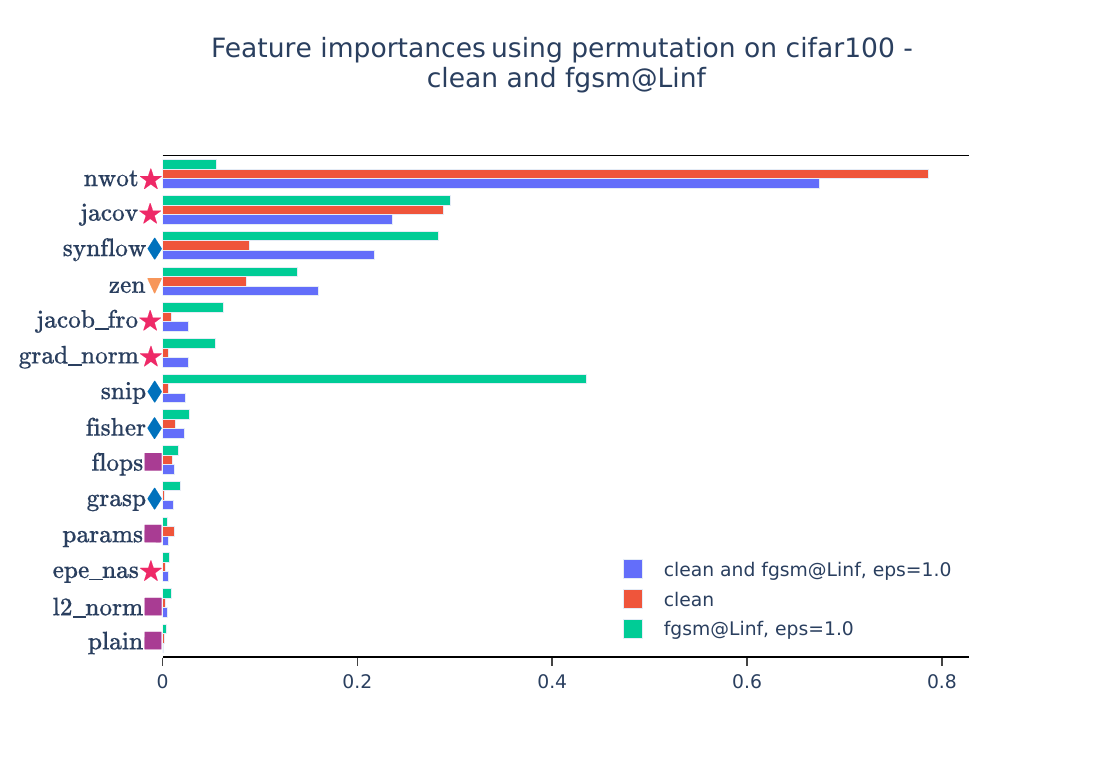}
\includegraphics[width=0.45\textwidth]{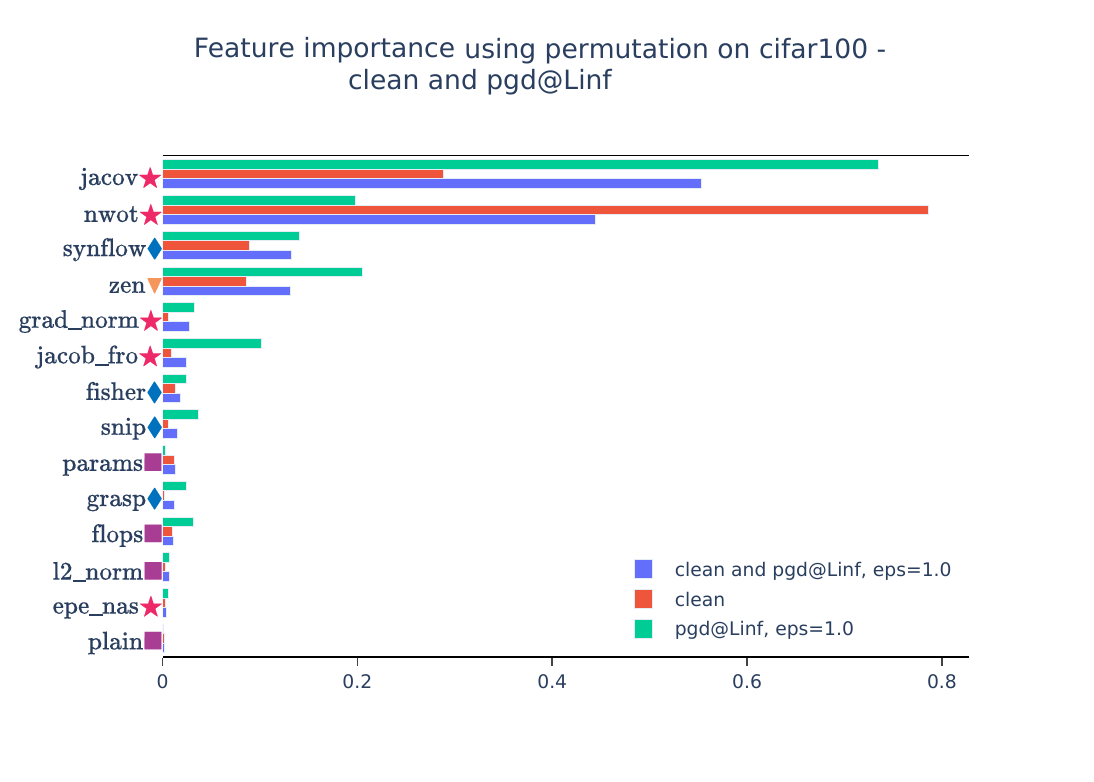}
\includegraphics[width=0.45\textwidth]{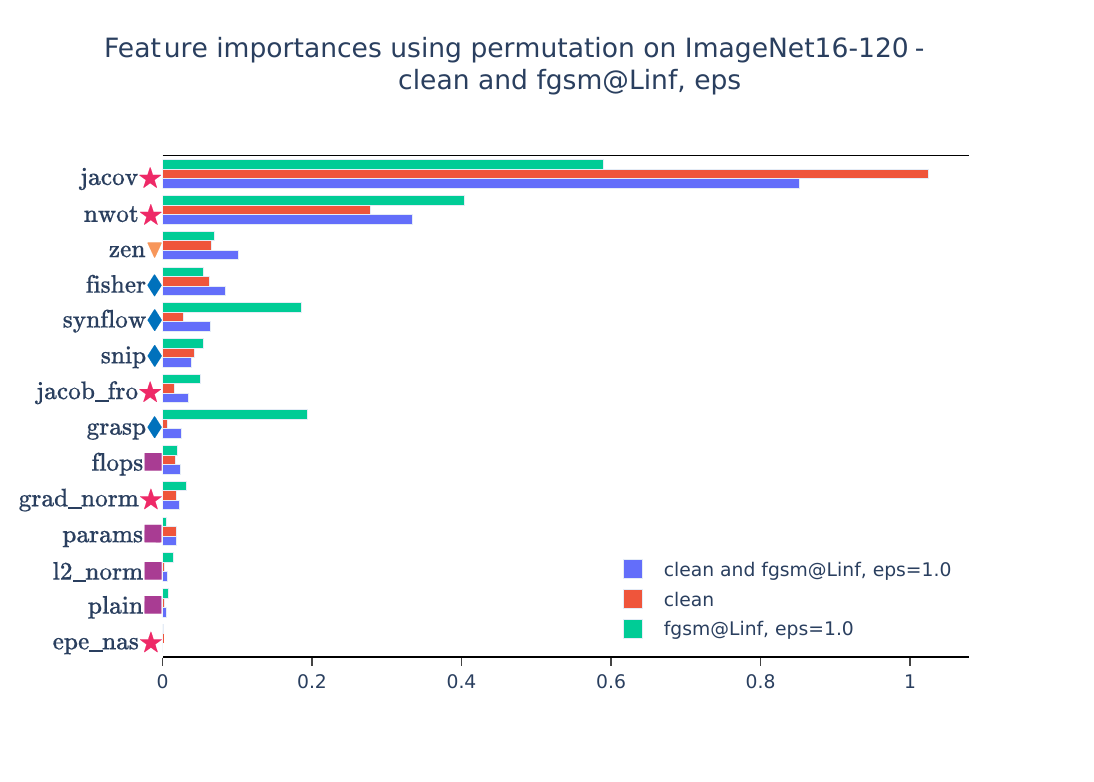}
\includegraphics[width=0.45\textwidth]{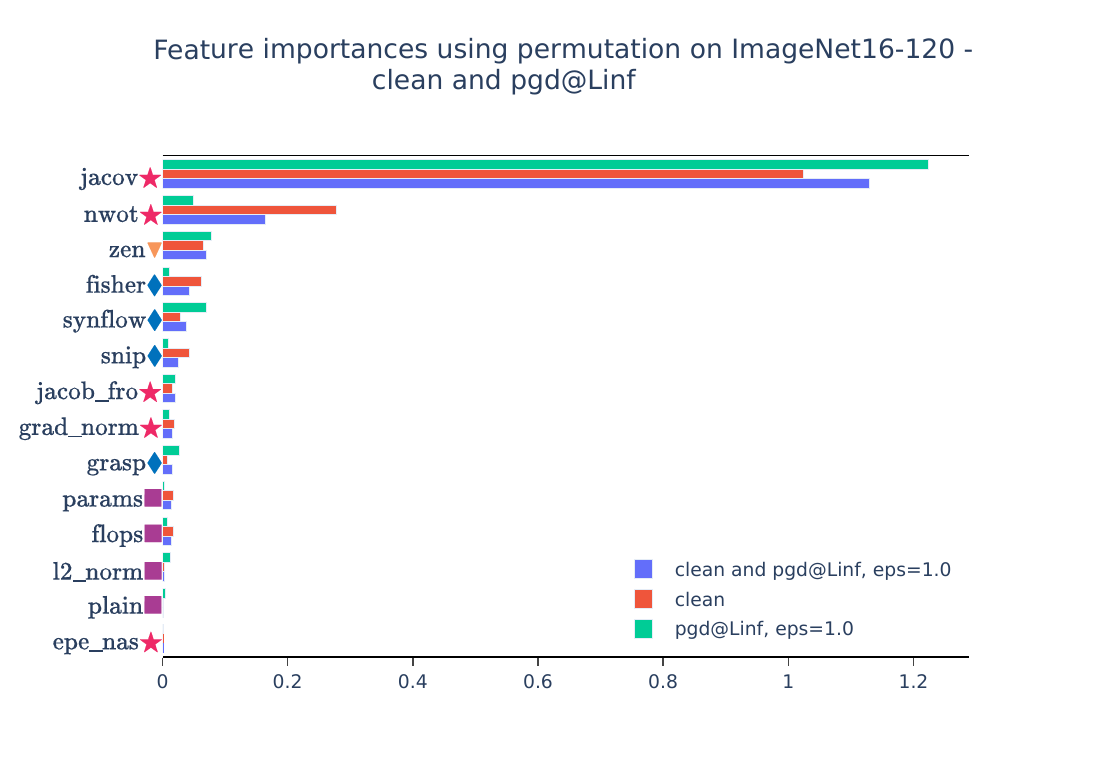}
\caption{Permutation importance of the random forest prediction model trained on $80\%$ of the data provided in \cite{Jung2023} with all zero-cost proxies as features and multi targets being clean test accuracy and different adversarial accuracies on \textbf{CIFAR-10} (\textbf{top}), CIFAR-100 (\textbf{middle}), and ImageNet16-120 (\textbf{bottom}).  
\label{fig1:pm_importance}}
\end{figure}

In this section we furthermore analyze the permutation-based feature importance of all considered zero-cost proxies on all three image datasets for the attacks FGSM and PGD with $\epsilon=1/255$ in Fig. \ref{fig1:pm_importance}.
In comparison to the feature importance from the main paper, we see here a more heterogeneous ZCP-type behavior. It is noticeable that the ZCP \texttt{zen} becomes here more important. 

\end{document}